\pdfoutput=1

\documentclass[11pt]{article}

\usepackage[]{acl}

\usepackage{times}
\usepackage{latexsym}

\usepackage[T1]{fontenc}

\usepackage[utf8]{inputenc}

\usepackage{microtype}

\usepackage{inconsolata}

\usepackage{graphicx}
\usepackage{amsmath}
\usepackage{adjustbox}
\usepackage{multirow}
\usepackage{enumitem}
\usepackage{amsfonts}   
\usepackage{booktabs}
\usepackage{pifont}
\newcommand{\cmark}{\checkmark}
\newcommand{\xmark}{\ding{55}}
\usepackage{amssymb}
\usepackage{float}

\usepackage[most]{tcolorbox}
\usepackage{lipsum} 
\usepackage{setspace}
\definecolor{lightgray}{gray}{0.8} 
\definecolor{pers}{RGB}{230,245,255} 
\definecolor{abn}{RGB}{255,245,230}  

\usepackage[table]{xcolor}
\usepackage{tcolorbox}
\usepackage{csquotes}
\usepackage{subcaption}
\usepackage{fancyhdr}

\title{\includegraphics[height=2em]{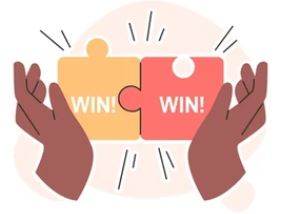} \textsc{prisma}: Preference-Reinforced Self-Training Approach for Interpretable Emotionally Intelligent Negotiation Dialogues}

\author{Prajwal Vijay Kajare$^{\ddagger}$\thanks{Authors are jointly first authors.}, Priyanshu Priya$^{\dagger\text{\textcolor{black}{*}}}$, Bikash Santra$^\ddagger$, \textbf{Asif Ekbal}$^\dagger$  \\
$^\dagger$Department of Computer Science and Engineering, Indian Institute of Technology Patna, India  \\
        $^\ddagger$School of Artificial Intelligence and Data Science, Indian Institute of Technology Jodhpur, India\\
        \texttt{$^\ddagger$\{p24ai0004,bikash}\}{\tt @iitj.ac.in}, \texttt{$^\dagger$\{priyanshu\_2021cs26,asif\}}{\tt @iitp.ac.in}}

\begin{document}
\maketitle
\begin{abstract}
Emotion plays a pivotal role in shaping negotiation outcomes, influencing trust, cooperation, and long-term relationships. Developing negotiation dialog systems that can recognize and respond strategically to emotions is, therefore, essential to create more effective human-centered interactions. Beyond generating emotionally appropriate responses, \textit{interpretability} - understanding how a system generates a particular emotion-aware response, is critical for fostering reliability and building rapport. Driven by these aspects, in this work, we introduce \textsc{prisma}, an interpretable emotionally intelligent negotiation dialogue system targeting two application domains, \textit{viz.} job interviews and resource allocation. To enable interpretability, we propose an Emotion-aware Negotiation Strategy-informed Chain-of-Thought (ENS-CoT) reasoning mechanism, which mimics human negotiation by perceiving, understanding, using, and managing emotions. Leveraging ENS-CoT, we curate two new datasets: \texttt{JobNego} (for job interview negotiation) and \texttt{ResNego} (for resource allocation negotiation). We then leverage these datasets to develop \textsc{prisma} by augmenting self-training with Direct Preference Optimization (DPO), guiding agents toward more accurate, interpretable, and emotionally appropriate negotiation responses. Automatic and human evaluation on \texttt{JobNego} and \texttt{ResNego} datasets demonstrate that \textsc{prisma} substantially enhances interpretability and generates appropriate emotion-aware responses, while improving overall negotiation effectiveness\footnote{Code and dataset are available at \\\url{https://github.com/kajareprajwal1143/PRISMA} and \url{https://ai-nlp-ml.github.io/resources.html}.}.
\end{abstract}

\section{Introduction}
\begin{figure}[!h]
    \centering
    \includegraphics[width=0.9\linewidth,trim={40 20 20 40},scale = 1.0]{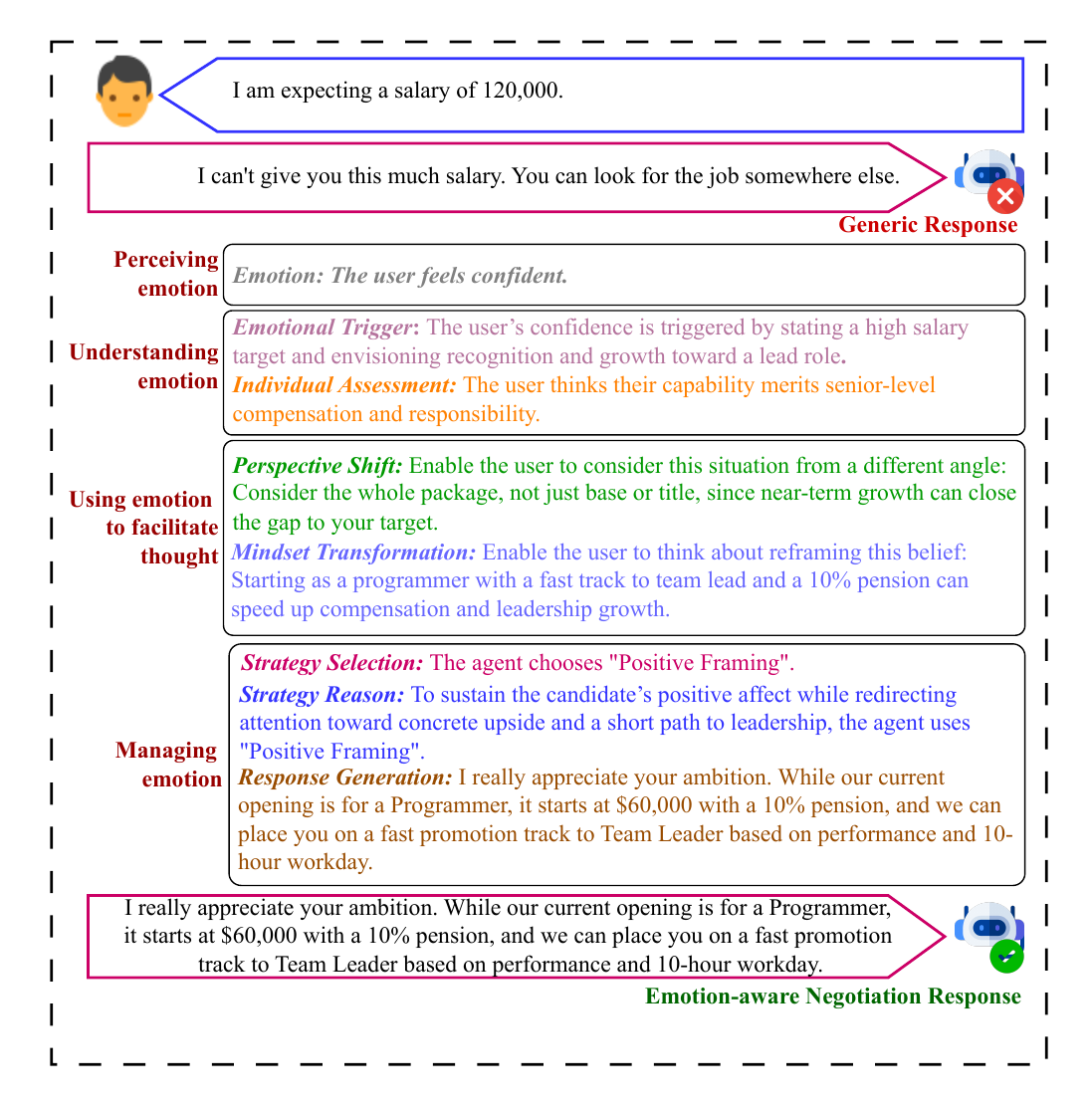}
    \caption{A dialogue snippet illustrating the role of emotional intelligence with interpretable ENS-CoT rationales in generating emotion-aware responses and enhancing negotiation outcomes.}
    \label{motivation}
\end{figure}

Negotiation is inherently a social process where parties seek to settle exchanges to satisfy their needs \cite{spector1977negotiation}. Traditionally, research has examined negotiation through economic modeling \cite{sebenius1992negotiation}, decision-making theory \cite{bazerman2000negotiation}, and social psychology \cite{rubin2013social}. However, these classical views often portray negotiators as purely rational agents, overlooking the pivotal role of emotions. Behavioral evidence shows that emotions systematically drive concession-making, perceived fairness, joint gains, and cooperation \cite{lawler1995structural,forgas1998feeling}. Thus, emotional intelligence (EI: the ability to perceive, understand, use, and manage emotions) \cite{goleman2005emotional}, is vital for developing human-like negotiation dialogue agents.

Although modern agents based on the large language models (LLMs) can reason via chain-of-thought (CoT) \cite{wei2022chain} and recognize emotions \cite{li2024enhancing}, they remain limited in strategically leveraging emotions during negotiation. They often generate static rather than adaptive emotion-aware strategies \cite{kroczek2025influence}, struggle to distinguish genuine from manipulated emotions \cite{huang2024personality}, and lack reasoning about emotional cause-and-effect dynamics \cite{bosse2013learning}. These limitations are critical in high-stakes negotiation settings like job interviews and resource allocation, where emotion directly influences strategic decisions (e.g., frustration may hasten concessions, while excitement can trigger overly ambitious offers), shaping negotiation outcomes. For such agents to be reliable, users must understand the rationale behind their emotionally-tuned responses \cite{dang2024explainable}. Hence, beyond emotional intelligence, a negotiation agent must also be interpretable; not merely generate responses, but also explain how it generates these responses to build rapport and foster credibility. 

Motivated by this need, we introduce \textsc{prisma}, an interpretable, emotionally intelligent negotiation dialogue system that first reasons via a CoT mechanism and then generates emotion-aware negotiation responses. To enable interpretable EI in \textsc{Prisma}, we draw inspiration from the Mayer–Salovey–Caruso Emotional Intelligence Test (MSCEIT) \cite{mayer2003measuring}, which encompasses four abilities: \textit{perceiving emotion} (identifying the user’s current emotion), \textit{understanding emotion} (explaining its cause and effect), \textit{using emotion to facilitate thought} (guiding reasoning through perspective shifting and mindset transformation), and \textit{managing emotion} (selecting and justifying an emotion-aware negotiation strategy, then realizing it in the response). As shown in Figure \ref{motivation}, unlike generic response that may frustrate the user, \textsc{prisma} interprets the user’s confidence for high salary, links it to aspirations for recognition and leadership, reframes priorities, and adopts a `Positive Framing' strategy emphasizing promotion and growth, leading to an emotionally attuned counteroffer to improve the negotiation outcome.

To advance this line of research, we introduce two novel negotiation dialogue datasets, \textit{viz.} \texttt{JobNego} and \texttt{ResNego}, generated using ChatGPT \cite{openai2024chatgpt} for job interview and resource allocation negotiations, respectively. These datasets are constructed through an interpretable emotion-aware dialogue generation process. We first design diverse, realistic negotiation scenarios, then, conditioned on scenarios, generate dialogues augmented with MSCEIT-inspired Emotion-aware Negotiation Strategy-informed CoT (ENS-CoT) rationales, where the agent first reasons to select appropriate emotion-aware strategies and then generates emotion-aware responses during negotiation. 

Curating large-scale, high-quality dialogue data with rationale annotations, however, is time- and resource-intensive. In contrast, unlabeled negotiation data can be synthesized at scale and leveraged via semi-supervised methods, such as self-training to improve LLM performance \cite{zelikman2022star,gulcehre2023reinforced}. Moreover, Direct Preference Optimization (DPO) has shown promise for aligning LLMs with preferred outputs \cite{rafailov2023direct,deng2024flow}. Building on these insights, we propose a preference-reinforced self-training approach that integrates self-training with DPO to develop \textsc{prisma}. It begins with supervised initialization to enable the LLM to generate ENS-CoT rationales and appropriate emotion-aware responses, followed by an iterative loop alternating DPO-based preference learning (aligning agents to preferred emotion-aware negotiation responses) and self-training (adding high-confidence ENS-CoT pseudo-labels for unlabeled contexts back into supervised fine-tuning). This integration optimizes step-by-step CoT reasoning, resulting in improved emotion-aware negotiation responses. Empirical evaluations indicate that the proposed approach enhances the interpretability and emotionally attuned negotiation behavior in \textsc{prisma}.  

In summary, the key contributions of this work are as follows: (i) Present \textsc{Prisma}, an interpretable, emotionally intelligent negotiation dialogue agent. To the best of our knowledge, this work pioneers interpretability modeling in emotion-aware negotiation conversations; (ii) Propose an Emotion-aware Negotiation Strategy-informed Chain-of-Thought (ENS-CoT) reasoning mechanism to enable interpretable emotional intelligence; (iii) Curate two novel datasets, \texttt{JobNego} (job interview negotiation) and \texttt{ResNego} (resource allocation negotiation), annotated with ENS-CoT rationales; and (iv) Introduce a preference-reinforced self-training approach integrating self-training with DPO, demonstrating its effectiveness in producing interpretable, emotion-aware negotiation responses.

\section{Related Work}
Negotiation has long been a central topic in psychology, economics, and affective computing \citep{carnevale1992negotiation,boothby2023embracing}. In recent years, automated negotiation dialogue systems have made significant progress \cite{lewis2017deal,zhou2019dynamic}, demonstrating that modeling negotiators' behavior \cite{he2018decoupling,zhang2020learning,chawla2022opponent}, strategies \cite{zhao2019rethinking,chawla2021casino,yamaguchi2021dialogue}, personalities \cite{yang2020improving,priya2024trip,priya2025we}, and politeness \cite{zhao2018sogo,priya2025genteel} can substantially enhance human-agent negotiation interactions. Emotions also play a critical role in shaping negotiation outcomes \cite{morris2000emotions,douglas2014attitude}, and a growing number of studies have begun integrating emotional cues into automated negotiation systems \cite{chawla2021towards,chawla2023towards,keskin2024negotiator}.

The advent of LLMs \cite{minaee2024large} has further spurred research in negotiation dialogue systems \cite{fu2023improving,gandhi2023strategic,abdelnabi2023llm,deng2023plug}. Emotional intelligence (awareness and management of emotions) \cite{maier2004emotional} and interpretability (ability to understand the reasoning behind the model's outputs) \cite{fuhrman2022review}, have become key considerations in designing reliable LLM-based agents \cite{li2023toward,yang2024xagents,feng2025emotionally,kasbouya2025emotional}. To support interpretability, Chain-of-Thought (CoT) prompting \cite{wei2022chain} has been developed, enabling LLMs to emulate human reasoning processes and produce more transparent, step-wise rationales. Building on this, a number of studies have explored the use of CoT prompting to enhance LLM reasoning across diverse applications \cite{zhang2022automatic,yang2023psycot}, including dialogue generation \cite{deng2023prompting}. 

Complementing CoT approaches, reinforcement learning (RL) techniques \cite{bai2022training} have been applied to further refine LLM reasoning \cite{trung2024reft,wang2025reinforcement}. Notably, Direct Preference Optimization (DPO) \cite{rafailov2023direct}, which aligns model outputs with human preferences, has shown promise in improving reasoning and output quality \cite{deng2024flow,lai2024step}. Additionally, semi-supervised methods \cite{scudder1965probability,amini2025self}, such as self-training \cite{amini2025self}, have gained traction for enhancing reasoning capabilities in LLMs, providing a scalable mechanism for leveraging unlabeled or weakly labeled data \cite{lee2024self,wu2024rethinking}.

Despite extensive research on modeling opponents' behavior, strategies, emotions, and politeness in developing negotiation dialogue agents, little research has examined the specific role of interpretable emotional intelligence within negotiation dialogues. Moreover, combining self-training with preference signals remains underexplored, especially in negotiation scenarios where interpretability and emotional intelligence are critical. To address these gaps, we introduce \textsc{prisma}, a preference-reinforced self-training approach that leverages Emotion-aware Negotiation Strategy-informed CoT (ENS-CoT) rationales to align responses with users' emotional states while ensuring interpretability during negotiation. \textsc{prisma} is designed to be scalable and adaptable across diverse application domains.

\section{Dataset}
We evaluate \textsc{prisma} on two newly created negotiation dialogue datasets: \texttt{JobNego} and \texttt{ResNego}. \texttt{JobNego} comprises job interview negotiations between a candidate (user) and an employer (agent), covering salary, working hours, promotion track, and related terms. \texttt{ResNego} includes resource-allocation negotiation dialogues in a camping setting, where parties bargain over food, water, firewood, and related supplies.

\subsection{Dataset Preparation}
Given the high cost of creating interpretable, emotionally intelligent negotiation dialogue datasets, where the agent must combine domain constraints and trade-offs with emotion-aware negotiation strategy selection, we adopt an interpretable and scalable dataset construction scheme. Specifically, we introduce an Emotion-aware Negotiation Strategy-informed Chain-of-Thought (ENS-CoT) reasoning mechanism to generate emotion-aware negotiation responses in an interpretable manner. ENS-CoT mimics human negotiation behavior by identifying, understanding, using, and managing emotions during negotiation through appropriate emotion-aware negotiation strategies. Using ENS-CoT, we curate two new interpretable emotion-aware negotiation dialogue datasets: \texttt{JobNego} and \texttt{ResNego} for job interview and resource allocation domains, respectively. The entire dataset construction involves three steps: (1) Negotiation Scenario Generation, (2) Emotion and Negotiation Strategy Formulation, and (3) ENS-CoT-enhanced Dialogue Generation.

\paragraph{(1) Negotiation Scenario Generation.} The scenarios summarize the negotiation setup, including task, roles, issues, constraints, and goals underlying each dialogue. The diverse and realistic scenarios are essential for generating high-quality negotiation dialogues. We, therefore, use NegoChat \cite{konovalov2016negochat} and CaSiNo \cite{chawla2021casino} datasets as seed corpora to generate the initial pool of job interview and resource allocation negotiation scenarios, respectively. For each dialogue, we prompt ChatGPT in a zero-shot setting to produce a brief scenario description. 

To identify an effective prompt, we design four candidate prompts (including the natural language instructions and target dialogues), and for each prompt, we generate 30 scenarios per domain. These generated scenarios are then manually evaluated by the three human annotators for scenario appropriateness on a scale of 1-3 (1-low, 2-moderate, 3-high)\footnote{The annotators hold Ph.D. in Linguistics with extensive expertise in negotiation, emotional intelligence, and dialogue design. They are compensated as per the institute norms.}. An inter-annotator Kappa agreement score ($\kappa$) \cite{mchugh2012interrater} reaches 0.85 (for job interview scenarios) and 0.88 (for resource allocation scenarios). The prompt generating the highest number of scenarios with a score of 3 for both domains is selected as the final prompt. This prompt is then used to prompt ChatGPT to generate scenarios for all the dialogues. The generated scenarios are further verified by the same annotators, achieving $\kappa = 0.88$ and $0.86$ among them for job interview and resource allocation scenarios, respectively. It yields the unique 105 and 1,030 job interview and resource allocation scenarios, respectively.

We expand this scenario pool by leveraging ChatGPT's in-context learning ability \cite{wang2023self}. We randomly sample three dialogue–scenario exemplars per domain from the initial pool and use them to prompt ChatGPT in a few-shot setup to generate additional scenarios. To ensure quality and diversity, we remove duplicates and filter out inadequate scenarios that lack proper information on negotiation setup or have incomplete sentences. This yields an additional 735 job interviews and 618 resource allocation scenarios, enriching the pool for subsequent dialogue generation. The scenario generation prompt template is given in the Appendix \ref{dataset_details}.

\paragraph{(2) Emotions and Emotion-aware Negotiation Strategies Formulation.} Emotions significantly influence negotiation dynamics and outcomes \cite{barry1996affect}. Users naturally express positive, negative, or neutral emotions during negotiation, and these emotions determine how negotiation unfolds \cite{overbeck2010feel}. In this work, we adopt 12 emotion categories from the negotiation literature \cite{barry1999tactical,van2008emotion}, including \textit{joy}, \textit{confidence}, \textit{positivity}, \textit{gratitude}, \textit{trust}, \textit{surprise}, \textit{anger}, \textit{disappointment}, \textit{frustration}, \textit{fear}, \textit{anxiety}, and \textit{neutral}. 

Effective negotiation depends on how the agent manages users' emotions \cite{van2022moving}. 
The dynamic nature of users’ emotions directly guides the agent's choice of strategy and, in turn, negotiation outcomes \cite{lawler1999bringing}. To effectively manage these emotions and drive negotiation toward mutually beneficial outcomes, we define 12 emotion-aware negotiation strategies grounded in negotiation emotion management principles \cite{yurtsever2004emotional}. These strategies are designed to either amplify positive emotions, mitigate negative ones, or maintain neutrality, and include \textit{savoring}, \textit{positive reinforcement}, \textit{expressing optimism}, \textit{cognitive reappraisal}, \textit{positive framing}, \textit{emotion diffusion}, \textit{expressive suppression}, \textit{active listening}, \textit{perspective-taking}, \textit{problem solving}, \textit{escalate assurance}, and \textit{no strategy}. Definitions of these strategies are given in Table \ref{strategy_def}. 

\begin{table*}[t]
\centering
\small
\begin{adjustbox}{max width=0.8\linewidth}
\begin{tabular}{p{3.2cm}|p{13cm}}
\hline
\textbf{Emotion-aware Negotiation Strategy} & \textbf{Definition} \\
\hline
\textbf{Savoring} & Actively appreciating and amplifying positive moments (e.g., shared successes, agreement points) to maintain a constructive climate and strengthen relationships to help achieve integrative outcomes. \\ \hline
\textbf{Positive Reinforcement} & Compliment and acknowledge constructive behavior or ideas to enhance positive affect, reinforce cooperation, and make progress toward mutual goals easier. \\ \hline
\textbf{Expressing Optimism} & Communicate a credible, positive outlook about reaching a mutually beneficial deal to encourage cooperative effort. \\ \hline
\textbf{Cognitive Reappraisal} & Reinterpret the situation to alter its emotional impact (e.g., view criticism as useful feedback) to reduce defensiveness and keep focus on objectives; linked to better outcomes.  \\ \hline
\textbf{Positive Framing} & Shift emphasis from potential losses to achievable gains to transform competitive stances into collaborative problem-solving. \\ \hline
\textbf{Emotion Diffusion} & Acknowledge heightened affect and de-escalate with calm, soft language and constructive addressing of issues to prevent conflict escalation.  \\ \hline
\textbf{Expressive Suppression} & Temporarily inhibit or mask one’s emotional display to maintain composure and avoid escalation in sensitive moments. \\ \hline
\textbf{Active Listening} & Attend to verbal and non-verbal cues, paraphrase, and validate concerns so the counterpart feels heard and understood; builds trust and diagnostic clarity. \\ \hline
\textbf{Perspective-Taking} & Deliberately adopt the counterpart’s viewpoint to infer their motives, constraints, and emotions, enabling more appropriate and empathetic responses. \\ \hline
\textbf{Problem Solving} & Collaboratively identify, analyze, and resolve issues to craft integrative solutions that meet mutual needs and goals. \\ \hline
\textbf{Escalate Assurance} & Address concerns with concrete guarantees, clarifications, and commitments (e.g., evidence, safeguards) to increase confidence and trust, stabilizing the interaction. \\ \hline
\textbf{No Strategy} & Refers to a neutral, task-focused response without explicit emotion management when affect is low/neutral or strategy use is unwarranted. \\ \hline
\end{tabular}
\end{adjustbox}
\caption{The definition of different emotion-aware negotiation strategies. Examples are given in Appendix \ref{dataset_details}.}

\label{strategy_def}
\end{table*}

\paragraph{(3) ENS-CoT-enhanced Dialogue Generation.} It involves following steps:

\textbf{(a) ENS-CoT Reasoning Chain Description.} To encode interpretable emotional intelligence in negotiation dialogue systems, we introduce the Emotion-aware Negotiation Strategy-informed CoT (ENS-CoT) reasoning mechanism, grounded in the MSCEIT's four abilities \cite{mayer2003measuring}: Perceiving emotion, Understanding emotion, Using emotion to facilitate thought, and Managing emotion. Perceiving emotion refers to the ability to identify user's current emotion, understanding emotion explains its causes via cognitive appraisal (emotional trigger and individual assessment), using emotion to facilitate thought specifies how emotion guides reasoning (via perspective shifting and mindset transformation), and managing emotion selects and justifies an emotion-aware negotiation strategy and realizes it in agent's response. An ENS-CoT instance is denoted as an octuple $\langle$EM, ET, IA, PS, MT, SS, SR, RG$\rangle$, described below:

\begin{itemize}[leftmargin=0.5em,itemsep=0pt, parsep=0pt, topsep=0pt, partopsep=0pt]
\setlength{\itemsep}{0pt}
\setlength{\parskip}{0pt}
\setlength{\parsep}{0pt}
    \item \textbf{Emotion (EM)}: User's expressed emotion.
    \item \textbf{Emotional Trigger (ET)}: Cause (external situation or internal thought) that triggers the emotion.
    \item \textbf{Individual Assessment (IA)}: Personal interpretation, evaluation, and internal response to the emotional trigger, based on the user’s past experiences, beliefs, expectations, and personal values.
    \item \textbf{Perspective Shift (PS)}: Adopting an alternate viewpoint to reduce reactivity and trade-offs.
    \item \textbf{Mindset transformation (MT)}: Reframing limiting beliefs into constructive priors for decision making.
    \item \textbf{Strategy Selection (SS)}: Choosing a pertinent emotion-aware negotiation strategy.
    \item \textbf{Strategy Reason (SR)}: Reasoning behind the selected strategy.
    \item \textbf{Response Generation (RG)}: Using the selected strategy to generate agent's utterance in an ongoing dialogue.
\end{itemize}

\textbf{(b) Designing Sample ENS-CoT-enhanced Dialogues.} We author exemplar dialogues conditioned on the generated scenarios, the user emotion inventory, and the agent’s emotion-aware negotiation strategy catalog across both domains. For each domain, we create 10 Wizard-of-Oz dialogues \cite{kelley1984iterative} with two annotators role-playing the user and the agent. The dialogues proceed turn-by-turn, grounded in the scenario: the user produces an utterance and expresses an emotion; the agent applies ENS-CoT to perceive, understand, use, and manage that emotion before responding. Every agent turn is annotated with an ENS-CoT rationale, ensuring that responses are emotionally attuned, strategically appropriate, and interpretable. `Guidelines for Designing Sample ENS-CoT-enhanced Dialogues’ are provided in Appendix \ref{dataset_details}. 

\textbf{(c) Dialogue Synthesis via Prompting.} We use the authored dialogue samples as few-shot exemplars to prompt ChatGPT to generate entire dialogue datasets. To finalize the prompt, we experiment with four manually designed prompts, including task instructions with the target negotiation scenario, and few-shot exemplars (three randomly selected negotiation scenarios and their corresponding dialogues with ENS-CoT annotations). For each prompt, we generate 15 dialogues, each for both the domains by prompting ChatGPT. These generated dialogues are evaluated by the same three annotators for emotional intelligence (EI) and negotiation quality (NQ) on a 1-3 scale (low to high) under the supervision of a subject matter expert with HR and sales backgrounds. We obtain $\kappa=0.81 \text{ and } 0.84$ for EI and NQ, respectively, indicating good quality negotiation dialogues. We select the prompt that generates the highest number of dialogues rated as 3 for both EI and NQ, and use it to construct the final datasets. The final dataset statistics are reported in Table \ref{dataset_statistics}. The dialogue synthesis prompt template, along with a detailed analysis of the joint emotion–strategy distribution and scenario-level topic diversity for both JobNego and ResNego, is provided in Appendix \ref{dataset_details}.

\renewcommand{\arraystretch}{1.1}
\begin{table}[!hbt]\Large
\begin{center}
\begin{adjustbox}{max width=\linewidth}
\begin{tabular}{l|ccc|ccc}
\hline
 & \multicolumn{3}{c|}{\textbf{\texttt{JobNego}}} & \multicolumn{3}{c}{\textbf{\texttt{ResNego}}} \\ \hline
 & \textbf{Train} & \textbf{Dev} & \textbf{Test} & \textbf{Train} & \textbf{Dev} & \textbf{Test} \\ \hline
\textbf{\textit{\# Dialogues}} & 504 & 168 & 168 & 988 & 330 & 330 \\
\textbf{\textit{\# Utterances}} & 8,124 & 2,196 & 2,172 & 13,695 & 3,314 & 3,178 \\
\textbf{\textit{Avg. Utterances/Dialogue}} & 16.12 & 13.07 & 12.93 & 13.86 & 10.04 & 9.63 \\ \hline
\end{tabular}
\end{adjustbox}
\caption{\textbf{\texttt{JobNego}} and \textbf{\texttt{ResNego}} dataset statistics.}
\label{dataset_statistics}
 \end{center}
\end{table}

\begin{figure*}[!h]
    \centering
    \includegraphics[height=6cm, width=0.7\linewidth,trim={20 15 15 20},scale = 1.0]{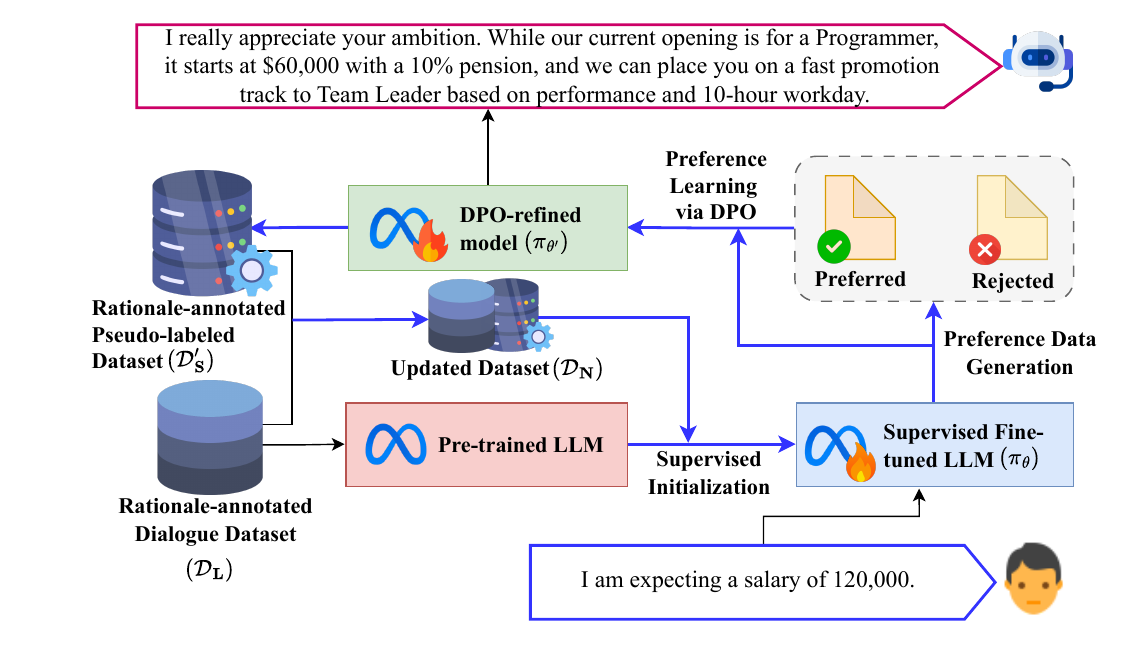}
    \caption{Architecture of the proposed \textsc{\textbf{p}}reference-\textsc{\textbf{r}}e\textsc{\textbf{i}}nforced \textsc{\textbf{s}}elf-training approach for an interpretable e\textsc{\textbf{m}}otionally intelligent negoti\textsc{\textbf{a}}tion dialogue system - \textsc{prisma}.}
    \label{architecture}
\end{figure*}
\subsection{Dataset Filtering and Quality Evaluation}
After generating the datasets, we evaluate each dialogue based on the following criteria: Emotional Intelligence (EI), Strategy Appropriateness (SA), Interpretability (IN), Fairness (F), Coherence (C), Naturalness (N), and Interestingness (I) (Descriptions are given in Appendix \ref{dataset_details}).  The same three annotators rate every dialogue on a 1-5 scale (low to high). 

We retain dialogues that receive scores $\ge 3$ on all the metrices. For \texttt{JobNego}, the retained dialogues attains average ratings of 4.62 (EI), 4.44 (SA), 4.27 (IN), 4.33 (F), 4.25 (C), 4.81 (N), and 4.34 (I), with $\kappa$ scores of 0.85, 0.83, 0.79, 0.82, 0.80, 0.82, and 0.78 for EI, SA, IN, F, C, N, and I, respectively. For \texttt{ResNego}, average ratings are 4.55 (EI), 4.39 (SA), 4.22 (IN), 4.40 (F), 4.20 (C), 4.77 (N), and 4.29 (I), with corresponding $\kappa$ scores of 0.83, 0.81, 0.80, 0.84, 0.78, 0.80, and 0.77. These results indicate that conversations are emotionally attuned, strategically appropriate, interpretable, fair, coherent, natural, and engaging across both domains.

\section{Methodology}
\label{sec:method}

The overall architecture of \textsc{prisma}, a \textsc{\textbf{p}}reference-\textsc{\textbf{r}}e\textsc{\textbf{i}}nforced \textsc{\textbf{s}}elf-training approach for an interpretable e\textsc{\textbf{m}}otionally intelligent negoti\textsc{\textbf{a}}tion dialogue system is depicted in Figure \ref{architecture}.

\noindent \textbf{Problem Formulation.} We present \textsc{prisma} that integrates emotion reasoning with strategy selection and response generation. The system is designed to (i) generate interpretable CoT rationales that explain how emotions influence negotiation strategies, and (ii) generate final responses that are emotionally attuned and strategically appropriate. Formally, given a dialogue context $\small c = {u_1, a_1 \dots, u_{t-1}, a_{t-1}}$, which is an alternating sequence of $(t-1)$ turns between the user $(u)$ and the dialogue agent $(a)$, and the user's target turn $(u_t)$, the negotiation agent \textsc{prisma} first generates an ENS-CoT rationale $r$. The agent's response $a_t = (y)$ is then generated conditioned on context $c$ and the selected strategy extracted from $r$. This ensures that the agent's response is emotionally and strategically appropriate, while also remaining interpretable, since the rationale explicitly shows how a specific strategy is chosen.

\noindent \textbf{Proposed Approach.} To develop \textsc{prisma}, we introduce a preference-reinforced self-training paradigm. This is a two-step training procedure that integrates self-training with DPO to guide the model toward accurate, interpretable, and emotionally appropriate responses.

\noindent \textbf{Step 1: Supervised Initialization.}
We begin by fine-tuning a pre-trained LLM, LLaMA-2-7b \citep{touvron2023llama}, on the labeled dataset \(\mathcal{D}_L=\{(c_i,r_i,y_i)\}_{i=1}^{L}\), where $L$ is the number of samples in $\mathcal{D}_L$. For the $i$-th input context $c_i$, let \(o_i=\mathrm{concat}(\langle\mathrm{R}\rangle,r_i,\langle/\mathrm{R}\rangle,\langle\mathrm{A}\rangle,y_i,\langle/\mathrm{A}\rangle)\) be the target output sequence. The supervised fine-tuned (SFT) model is obtained by \(\pi_\theta\) by minimizing the token-level negative log-likelihood \(\mathcal{L}_{\mathrm{SI}}(\theta)\):
\setlength{\abovedisplayskip}{6pt}
\setlength{\belowdisplayskip}{6pt}
\setlength{\abovedisplayshortskip}{4pt}
\setlength{\belowdisplayshortskip}{4pt}
\begin{equation}
\small
\mathcal{L}_{\mathrm{SI}}(\theta)
= -\mathbb{E}_{(c,r,y)\sim\mathcal{D}_L}
\left[ \sum_{m=1}^{M}
\log \pi_\theta\!\big(o_m \,\big|\, c, o_{1:m-1}\big) \right]
\end{equation}
where \(M\) denotes number of tokens in rationale+response sequence, and $o_m$ represents the $m$-th token in $o$. This step equips the model with a seed ability to generate ENS-CoT rationales and emotion-aware responses during negotiation.

\noindent \textbf{Step 2: Preference-Reinforced Self-Training.}
After initialization, we iteratively refine the model through two alternating phases:

\textbf{(a) Preference Learning via DPO.} For some context $c$ from the unlabeled dataset $\mathcal{D}_U=\{(c_i,y_i)\}_{i=1}^{U}$ ($U$ denotes the number of instances in $\mathcal{D}_U$), we sample multiple rationales $\hat{r}$ from $\pi_\theta$. We then build a preference dataset $\mathcal{D}_P = \{(c_i,r_i^+,r_i^-)\}_{i=1}^P$ consisting of preferred and rejected completions, where $r_i^+$ is the preferred completion, $r_i^-$ is the rejected completion, and $P$ is the size of the preference dataset. A rationale is marked \emph{preferred} if its corresponding response has cosine similarity to the ground-truth response $y_i$ exceeding a threshold $\tau_1$. It is marked \emph{rejected} if the similarity falls below a threshold $\tau_2$\footnote{\label{fn:thresholds}All thresholds are chosen empirically based on development-set similarity distributions, fixed globally across experiments, and verified for robustness via sensitivity analysis, as detailed in Appendix \ref{appendix_experiment}.}. 
We then update $\pi_\theta$ by minimizing the DPO loss $\mathcal{L}_{\text{PL}}(\theta')$ on $\mathcal{D}_P$, yielding DPO-refined model $\pi_{\theta'}$:
\setlength{\abovedisplayskip}{6pt}
\setlength{\belowdisplayskip}{6pt}
\setlength{\abovedisplayshortskip}{4pt}
\setlength{\belowdisplayshortskip}{4pt}
\begin{equation}
\small
\begin{aligned}
\mathcal{L}_{\text{PL}}(\theta')
= -\mathbb{E}_{(c,r^{+},r^{-})\sim\mathcal{D}_P} & \\
\left[
\log \sigma\!\left(
\beta\!\left(
\log \frac{\pi_{\theta'}(r^{+}\mid c)}{\pi_{\theta}(r^{+}\mid c)}
-
\log \frac{\pi_{\theta'}(r^{-}\mid c)}{\pi_{\theta}(r^{-}\mid c)}
\right)
\right)
\right]
\end{aligned}
\end{equation}

where, $\beta$ denotes the coefficient controlling the $\pi_{\theta'}$'s deviation from $\pi_{\theta}$.

\textbf{(b) Supervised Fine-Tuning with Updated Data.} Using DPO-refined policy $\pi_{\theta'}$, we generate rationales for context $c \in \mathcal{D}_U$. This forms a pseudo-labeled set $\mathcal{D}_S=\{(c,\hat{r},y)\mid (c, y)\sim\mathcal{D}_U,\; \hat{r}\sim\pi_{\theta'}(\cdot\mid c)\}$ for next round of supervised fine-tuning. We retain rationales whose responses have cosine similarity to the corresponding ground-truth responses exceeding $\tau_3$, and remove duplicates. This yields a clean dataset $\mathcal{D}'_S\subset\mathcal{D}_S$.
The model is then trained in a supervised way on the updated dataset, $\mathcal{D}_N=\mathcal{D}_L\cup\mathcal{D}'_S$. To mitigate the risk of overfitting and enable continual learning, we reinitialize training from the pre-trained LLaMA-2-7b model once the dataset is updated. We do this rather than continuing to fine-tune the SFT model $\pi_{\theta}$, following prior works \cite{zelikman2022star,singh2023beyond}. 

Step 2 repeats until convergence or the iteration limit, resulting in a negotiation agent that effectively identifies and interprets emotions, selects an appropriate emotion-aware negotiation strategy, and grounds its responses in the chosen strategy.

\begin{table*}[!t]
\centering
\begin{adjustbox}{max width=0.7\linewidth}
\begin{tabular}{lccccccc}
\toprule
\textbf{Models} & \textbf{PPL} $\downarrow$ & \textbf{B-4} $\uparrow$ & \textbf{D-3} $\uparrow$ & \textbf{BS-F1} $\uparrow$ & \textbf{R-LEN} $\uparrow$ & \textbf{EA} $\uparrow$ & \textbf{ENSC} $\uparrow$ \\
\hline
\multicolumn{8}{c}{\textbf{\texttt{JobNego}}} \\
\hline
Llama-2-7B-SFT & 6.984 & 5.806 & 0.266 & 0.886 & 20.917 & 0.572 & 0.751 \\
Llama-3.2-3B-SFT & 7.318 & 4.611 & 0.271 & 0.888 & 21.204 & 0.579 & 0.756 \\
Emotion-aware SFT (Llama-2-7b) & 6.813 & 6.312 & 0.288 & 0.896 & 23.467 & 0.712 & 0.781 \\
ICL-AIF (GPT-3.5) & 9.612 & 5.321 & 0.198 & 0.852 & 12.904 & 0.462 & 0.641 \\
AnE & 7.623 & 5.679 & 0.253 & 0.879 & 19.102 & 0.611 & 0.737 \\
Standard Prompting (GPT-5) & 8.937 & 5.447 & 0.209 & 0.858 & 14.906 & 0.486 & 0.667 \\
Proactive Prompting (GPT-5) & 7.411 & 5.732 & 0.259 & 0.883 & 19.634 & 0.566 & 0.744 \\
ProCoT (GPT-5) & 7.842 & 5.618 & 0.247 & 0.875 & 18.439 & 0.642 & 0.728 \\
PPDPP & 7.566 & 5.637 & 0.251 & 0.878 & 19.221 & 0.744 & 0.733 \\
RLHF & 6.489 & 6.277 & 0.300 & 0.902 & 24.612 & 0.814 & 0.798 \\\hline
\rowcolor{lightgray}
\textsc{\textbf{prisma}} & \textbf{5.196} & \textbf{7.119} & \textbf{0.322} & \textbf{0.912} & \textbf{27.108} & \textbf{0.842} & \textbf{0.823} \\
\hline
\multicolumn{8}{c}{\textbf{\texttt{ResNego}}} \\
\hline
Llama-2-7b-SFT & 6.203 & 6.194 & 0.292 & 0.899 & 24.512 & 0.607 & 0.789 \\
Llama-3.2-3b-SFT & 6.589 & 5.190 & 0.297 & 0.901 & 24.933 & 0.612 & 0.794 \\
Emotion-aware SFT (Llama-2-7b) & 5.982 & 7.208 & 0.313 & 0.909 & 27.521 & 0.755 & 0.819 \\
ICL-AIF (GPT-5) & 8.537 & 6.143 & 0.221 & 0.869 & 15.318 & 0.501 & 0.682 \\
AnE & 6.842 & 6.183 & 0.283 & 0.894 & 23.102 & 0.644 & 0.775 \\
Standard Prompting (GPT-5) & 8.011 & 6.154 & 0.236 & 0.874 & 17.842 & 0.523 & 0.706 \\
Proactive Prompting (GPT-5) & 6.702 & 6.188 & 0.289 & 0.897 & 23.987 & 0.601 & 0.783 \\
ProCoT (GPT-5) & 6.993 & 6.179 & 0.277 & 0.891 & 22.311 & 0.683 & 0.767 \\
PPDPP & 6.912 & 6.181 & 0.279 & 0.892 & 22.987 & 0.787 & 0.771 \\
RLHF  & 5.741 & 7.218 & 0.324 & 0.913 & 29.004 & 0.852 & 0.836 \\\hline
\rowcolor{lightgray}
\textsc{\textbf{prisma}} & \textbf{4.462} & \textbf{8.231} & \textbf{0.341} & \textbf{0.922} & \textbf{32.137} & \textbf{0.874} & \textbf{0.861} \\
\bottomrule
\end{tabular}
\end{adjustbox}
\caption{Automatic evaluation results. Results are statistically significant at 5\% significance level based on t-test \cite{welch1947generalization}.}
\label{tab:jobnego-resnego}
\end{table*}


\begin{table}[!hbt]
\begin{center}
\begin{adjustbox}{max width=\linewidth}
\begin{tabular}{lccccccc} \toprule
\textbf{Models} & \textbf{F} & \textbf{C} & \textbf{E} & \textbf{EA} & \textbf{ENSC} & \textbf{BE} & \textbf{OF} \\ \hline
\multicolumn{8}{c}{\textbf{\texttt{JobNego}}} \\ \hline
Llama-2-7B-SFT                 & 3.65 & 3.50 & 3.40 & 3.45 & 3.42 & 3.30 & 3.28 \\
Emotion-aware SFT & 4.10 & 3.98 & 4.05 & 4.12 & 4.00 & 3.95 & 3.90 \\
ProCoT            & 3.88 & 3.75 & 3.70 & 3.82 & 3.68 & 3.65 & 3.60 \\
PPDPP                          & 4.00 & 3.90 & 3.85 & 3.98 & 3.92 & 3.80 & 3.78 \\
RLHF                           & 4.35 & 4.22 & 4.30 & 4.38 & 4.20 & 4.18 & 4.12 \\\hline
\rowcolor{lightgray}
\textbf{\textsc{prisma}}& \textbf{4.52} & \textbf{4.40} & \textbf{4.48} & \textbf{4.55} & \textbf{4.48} & \textbf{4.36} & \textbf{4.30} \\
\hline
\multicolumn{8}{c}{\textbf{\texttt{ResNego}}} \\
\hline
Llama-2-7B-SFT                 & 3.78 & 3.65 & 3.55 & 3.62 & 3.60 & 3.48 & 3.45 \\
Emotion-aware SFT & 4.24 & 4.12 & 4.18 & 4.26 & 4.18 & 4.05 & 4.00 \\
ProCoT           & 4.02 & 3.90 & 3.86 & 3.98 & 3.92 & 3.85 & 3.80 \\
PPDPP                          & 4.15 & 4.05 & 4.00 & 4.12 & 4.08 & 3.95 & 3.90 \\
RLHF (PPO)                     & 4.47 & 4.35 & 4.40 & 4.51 & 4.42 & 4.31 & 4.24 \\\hline
\rowcolor{lightgray}
\textbf{\textsc{prisma}}& \textbf{4.68} & \textbf{4.56} & \textbf{4.62} & \textbf{4.70} & \textbf{4.64} & \textbf{4.50} & \textbf{4.45} \\
\bottomrule
\end{tabular}
\end{adjustbox}
\caption{Human evaluation results. Results are statistically significant at 5\% significance level based on t-test \cite{welch1947generalization}. All metrics are rated on a scale of 1 to 5.}
\label{humanres}
 \end{center}
\end{table}


\begin{table*}[hbt!]
\centering
\footnotesize
\begin{adjustbox}{max width=0.9\linewidth}
\begin{tabular}{l|cccccccc|ccccccc}
\toprule
& \multicolumn{8}{c|}{\textbf{Setting}} & \multicolumn{7}{c}{\textbf{Results}} \\
\cline{2-9}\cline{10-16}
\textbf{ID} & \textbf{EM} & \textbf{ET} & \textbf{IA} & \textbf{PS} & \textbf{MT} & \textbf{SS} & \textbf{SR} & \textbf{RG} & \textbf{PPL} $\downarrow$ & \textbf{B-4} $\uparrow$ & \textbf{D-3} $\uparrow$ & \textbf{BS-F1} $\uparrow$ & \textbf{R-LEN} $\uparrow$ & \textbf{EA} $\uparrow$ & \textbf{ENSC} $\uparrow$ \\
\hline
\multicolumn{16}{c}{\textbf{\texttt{JobNego}}} \\\hline
\rowcolor{lightgray}
0 & \cmark & \cmark & \cmark & \cmark & \cmark & \cmark & \cmark & \cmark & \textbf{5.196} & \textbf{7.119} & \textbf{0.322} & \textbf{0.912} & \textbf{27.108} &\textbf{ 0.842} & \textbf{0.823} \\\hline
1 & \cmark & \xmark & \xmark & \cmark & \cmark & \cmark & \cmark & \cmark & 6.024 & 6.486 & 0.309 & 0.904 & 24.816 & 0.816 & 0.788 \\
2 & \cmark & \cmark & \cmark & \xmark & \xmark & \cmark & \cmark & \cmark & 6.052 & 6.453 & 0.312 & 0.904 & 25.093 & 0.828 & 0.792 \\
3 & \cmark & \cmark & \cmark & \cmark & \cmark & \xmark & \xmark & \cmark & 6.081 & 5.352 & 0.310 & 0.898 & 25.612 & 0.834 & 0.741 \\
4 & \xmark & \xmark & \xmark & \xmark & \xmark & \cmark & \cmark & \cmark & 5.951 & 6.903 & 0.298 & 0.896 & 24.039 & 0.782 & 0.702 \\
5 & \xmark & \xmark & \xmark & \xmark & \xmark & \cmark & \xmark & \cmark & 5.879 & 5.204 & 0.292 & 0.890 & 23.521 & 0.731 & 0.648 \\
\hline
\multicolumn{16}{c}{\textbf{\texttt{ResNego}}} \\\hline
\rowcolor{lightgray}
0 & \cmark & \cmark & \cmark & \cmark & \cmark & \cmark & \cmark & \cmark & \textbf{4.462} & \textbf{8.231} & \textbf{0.341} & \textbf{0.922} & \textbf{32.137} & \textbf{0.874} & \textbf{0.861} \\\hline
1 & \cmark & \xmark & \xmark & \cmark & \cmark & \cmark & \cmark & \cmark & 4.339 & 7.617 & 0.326 & 0.912 & 29.842 & 0.853 & 0.829 \\
2 & \cmark & \cmark & \cmark & \xmark & \xmark & \cmark & \cmark & \cmark & 5.392 & 6.742 & 0.332 & 0.914 & 30.108 & 0.862 & 0.834 \\
3 & \cmark & \cmark & \cmark & \cmark & \cmark & \xmark & \xmark & \cmark & 5.409 & 6.664 & 0.333 & 0.909 & 30.573 & 0.868 & 0.812 \\
4 & \xmark & \xmark & \xmark & \xmark & \xmark & \cmark & \cmark & \cmark & 5.271 & 7.118 & 0.318 & 0.905 & 28.721 & 0.821 & 0.768 \\
5 & \xmark & \xmark & \xmark & \xmark & \xmark & \cmark & \xmark & \cmark & 5.205 & 7.431 & 0.311 & 0.898 & 28.106 & 0.783 & 0.722 \\
\bottomrule
\end{tabular}
\end{adjustbox}
\caption{Ablation study to analyze the impact of different components of ENS-CoT rationales in \textsc{prisma}.}
\label{ablation_rationales}
\end{table*}

\section{Experiments}
We compare \textsc{prisma} with 10 baselines: Llama-2-7b-SFT, Llama-3.2-3b-SFT \cite{touvron2023llama}, Emotion-aware SFT (Llama-2-7b), ICL-AIF \cite{fu2023improving}, AnE \cite{zhang2023ask}, Standard, Proactive, ProCoT (GPT-5) \cite{deng2023prompting,openai2025gpt5}, PPDPP \cite{deng2024plug}, RLHF \cite{ouyang2022training}. We evaluate \textsc{mediator} with both automatic and human judgments. For automatic evaluation, responses' general quality is measured by Perplexity (PPL) \cite{brown1992estimate}, BLEU-4 (B-4) \cite{papineni2002bleu}, BERTScore-F1 (BS-F1) \cite{zhang2019bertscore}, Distinct-3 (D-3) \cite{li2015diversity}, and Response Length (R-LEN); task success is assessed with Emotion Appropriateness (EA): evaluates the appropriateness of users' perceived emotions, and Emotion-aware Negotiation Strategy Consistency (ENSC): assesses the consistency of responses with intended strategies. For human evaluation, to assess the general quality of responses, we use Fluency (F), Coherence (C), and Engagingness (E); and to evaluate for task success, we mirror automatic metrics - EA and ENSC, and also report Bargaining Efficacy (BE) and Outcome Fairness (OF) \cite{ahmad2023ina}. Appendix \ref{appendix_experiment} provides `Implementation Details', `Baselines Details' and `Evaluation Metrics Details'. 

\section{Results and Analysis}
\noindent \textbf{Automatic Evaluation.} Table \ref{tab:jobnego-resnego} presents the automatic evaluation results for \textsc{prisma} and baselines on \texttt{JobNego} and \texttt{ResNego} datasets. Across both datasets, \textsc{prisma} consistently achieves superior dialogue generation quality, emotional intelligence, and negotiation performance. In terms of lexical and semantic metrics, it records the highest B-4, D-3, and BS-F1 scores. These results reflect enhanced fluency, diversity, and semantic alignment. Furthermore, the model generates longer responses (R-LEN) while maintaining emotional appropriateness (EA) and emotion-aware negotiation strategy consistency (ENSC). Specifically, on \texttt{JobNego}, \textsc{prisma} outperforms the strongest baseline, RLHF. It achieves significant gains of 13.4\%, 7.3\%, 1.1\%, 10.1\%, 3.4\%, and 3.1\% in B-4, D-3, BS-F1, R-LEN, EA, and ENSC, respectively. On \texttt{ResNego}, \textsc{prisma} achieves similar performance improvement. Compared to RLHF, it records gains of 14.0\%, 5.2\%, 1.0\%, 10.8\%, 2.6\%, and 3.0\% in B-4, D-3, BS-F1, R-LEN, EA, and ENSC, respectively. These improvements show that \textsc{prisma} generates linguistically richer responses. It also better captures emotional nuances and maintains consistent negotiation strategies. The performance gains could be attributed to the interpretability enforced through ENS-CoT rationales during preference-reinforced self-training, where modeling emotion-strategy reasoning sequences enables transparent, emotionally intelligent, and coherent negotiation behavior.

The baseline model, RLHF, tends to generate less diverse and less coherent emotion-aware responses during negotiation. Its inability to follow the complete reasoning chain might result in limited awareness and inconsistent understanding of the user's emotional states. Emotion-aware SFT performs well in emotional understanding. However, it tends to generate shorter, less strategic responses, thereby limiting negotiation effectiveness. 
ProCoT and PPDPP improve contextual control, but often underperform in emotional alignment and response diversity. In contrast, \textsc{prisma} balances emotional intelligence and negotiation efficiency, enhancing both the linguistic and affective aspects of dialogues. The consistent gains across datasets highlight its robustness and the effectiveness of the interpretable ENS-CoT reasoning mechanism.

\noindent \textbf{Human Evaluation.} Table \ref{humanres} presents human evaluation results for \textsc{prisma} and baselines. Given the high cost of manual evaluation, we compare \textsc{prisma} against Llama-2-7B-SFT, Emotion-aware SFT (Llama-2-7b), ProCoT (GPT-5-mini), PPDPP, and RLHF. On \texttt{JobNego}, \textsc{prisma} achieves average scores of 4.52, 4.40, 4.48, 4.55, 4.48, 4.36, and 4.30 for F, C, E, EA, ENSC, BE, and OF, respectively, showing consistent improvements of +0.17, +0.18, +0.18, +0.17, +0.28, +0.18, and +0.18 points over the RLHF. A similar trend is observed on \texttt{ResNego}. \textsc{prisma} outperforms RLHF by +0.21, +0.21, +0.22, +0.19, +0.22, +0.19, and +0.21 points on these metrics. These improvements highlight \textsc{prisma}’s ability to generate emotionally intelligent and strategically coherent negotiation dialogues. The higher F, C, and E scores demonstrate \textsc{prisma}’s ability to produce fluent, coherent, and engaging negotiation dialogues that sustain natural conversational flow. Besides, the superior EA, ENSC, BE, and OF scores highlight its effectiveness in maintaining emotional awareness, strategic consistency, and balanced outcome-focused negotiation behavior. These results indicate that interpretability strengthens emotional alignment and supports mutually beneficial negotiation outcomes. 

\noindent \textbf{Ablation Study.}To understand how each component of the ENS-CoT rationales affects model performance, we conduct an ablation study to investigate the impact of different components of the ENS-CoT rationales. Table \ref{ablation_rationales} reports results across different configurations in which one or more components are removed. In the first ablated model, we remove the `understanding emotion' components, specifically ET and IA. The second ablated model omits the `using emotion to facilitate thought' components, namely PS and MT. In the third ablated model, we eliminate the `managing emotion' components, which include SS and SR. Finally, the fourth ablated model excludes the components EM, ET, IA, PS, and MT, and the fifth ablated model removes the components EM, ET, IA, PS, MT, and SR. The removal of any rationale component adversely affects various critical aspects of dialogue quality, such as the appropriateness of emotions, the effectiveness of negotiations, and the language quality of responses. This is evidenced by reduced scores on metrics such as EA, ENSC, B-4, D-3, BS-F1, and R-LEN in both datasets. 

In particular, when we remove $\langle$ET, IA$\rangle$ or $\langle$PS, MT$\rangle$, B-4 increases slightly, but D-3 and BS-F1 drop, and responses become shorter. This shows that abilities like `understanding emotion' and `using emotion to facilitate thought' contribute substantive and varied content rather than n-gram overlap. Further, ablating $\langle$SS, SR$\rangle$ causes the largest decline in ENSC, indicating that explicit strategy choice and its justification are central to keeping the response aligned with the users' intended emotions. Also, excluding the `perceiving emotion' $\langle$EM$\rangle$ component along with the other components reduces EA, ENSC, D-3, BS-F1, and R-LEN. This pattern is consistent with simpler, more templated outputs that match surface forms but lose strategic consistency and emotional awareness. In general, these findings underscore the critical importance of each component in improving interpretability and encouraging the generation of emotion-aware responses within negotiation dialogues. 

\noindent \textbf{Additional Analysis.} We present more analyses: (1) Impact of Supervised Initialization, Self-Training, and DPO, (2) Impact of iterative training in \textsc{prisma}, and (3) Case Study in Appendix \ref{analysis_appendix}.

\section{Conclusion}
We presented \textsc{prisma}, an interpretable, emotionally intelligent negotiation dialogue agent designed to enhance negotiation outcomes in job interview and resource allocation domains. In this regard, we proposed an Emotion-aware Negotiation Strategy-informed CoT (ENS-CoT) reasoning mechanism that links perceived emotions to strategy selection and response generation. Using ENS-CoT, we curated \texttt{JobNego} and \texttt{ResNego} datasets. \textsc{prisma} integrates self-training with DPO through a preference-reinforced self-training framework, optimizing step-by-step CoT reasoning for emotion-aware negotiation responses. Extensive evaluation show that \textsc{prisma} achieves superior emotion appropriateness, strategy consistency, and negotiation effectiveness with improved interpretability.


\section*{Limitations}
The prompting-based data creation requires substantial compute and access to LLMs, and inherits both model biases and the variability of prompt-based generation. Although the resulting dialogues are generally fair and natural, quality remains bounded by the base model. The stronger controls beyond scenario prompts (e.g., constraint-based decoding, verifier models, or weak supervision) could further improve fidelity. Ensuring interpretability with ENS-CoT rationales also necessitates human verification, which constrains scale; realism is limited by assumptions of fluent interlocutors, coherent turn flow, and the absence of exogenous disruptions. Future work should automate correction/verification, broaden scenario/diversity coverage (culture, persona, etc.), and explore multi-emotion and multi-strategy annotations to better reflect real negotiation practice.

We also observe that \textsc{prisma} can occasionally produce prescriptive responses rather than fully collaborative ones, typically due to insufficient reasoning about user-specific constraints or implicit social norms (e.g., workplace expectations). Although human evaluations show that \textsc{prisma} generally behaves collaboratively, this limitation highlights an important direction for future work, which is, guiding ENS-CoT generation and preference learning toward explicitly collaborative problem-solving, emphasizing alternatives, phased plans, trade-offs, and shared decision-making rather than fixed prescriptions.

Further, the proposed negotiation dialogue agent, \textsc{prisma}, builds on self-training, but we do not systematically address unlabeled data usage. Preference construction for DPO depends on rubrics and thresholds that may encode hidden biases, and pseudo-labels can amplify early errors in \textsc{prisma}. Moreover, the evaluation is restricted to two bilateral, text-only domains (job interviews and camping resource allocation) primarily due to a lack of rationale-annotated negotiation dialogue dataset. Finally, several automatic proxies (e.g., similarity) do not substitute for user-centric outcomes; controlled user studies and online evaluations are an important next step to quantify trade-offs among interpretability, task success, and user experience.

\section*{Ethics Statement}
Interpretability is central for negotiation dialogue systems, especially when modeling emotional intelligence. Therefore, this work has been reviewed and approved by our Institutional Review Board (IRB). First, we treat dataset curation and distribution with care. All data sources are used in accordance with their licenses. We screen and filter generations with the help of linguistics and subject matter experts to reduce biased, offensive, insensitive, or inappropriate content to avoid potential unpredictable ethical harm. All human annotators and evaluators are paid according to the institute's norms. To ensure responsible data use, access will be granted only after the completion and signing of an agreement stipulating that the data will be used solely for research purposes.

Second, negotiation involves strategic influence, so the design and deployment of \textsc{prisma} must also adhere to strict ethical standards. \textsc{prisma} is explicitly optimized for collaborative, non–zero-sum outcomes: its emotion-aware strategies seek mutually beneficial agreements rather than one-sided wins. Crucially, user autonomy is preserved at all times, that is, users can decline proposals that do not meet their needs, and the system must not pressure, coerce, or manipulate. \textsc{prisma}'s ENS-CoT rationales are exposed to support transparency and human oversight, enabling users to understand how the dialogue agent decides a particular strategy.


\bibliography{manuscript}

@article{barry1996affect,
  title={Affect in dyadic negotiation: A model and propositions},
  author={Barry, Bruce and Oliver, Richard L},
  journal={Organizational Behavior and Human Decision Processes},
  volume={67},
  number={2},
  pages={127--143},
  year={1996},
  publisher={Elsevier}
}

@article{overbeck2010feel,
  title={I feel, therefore you act: Intrapersonal and interpersonal effects of emotion on negotiation as a function of social power},
  author={Overbeck, Jennifer R and Neale, Margaret A and Govan, Cassandra L},
  journal={Organizational Behavior and Human Decision Processes},
  volume={112},
  number={2},
  pages={126--139},
  year={2010},
  publisher={Elsevier}
}

@article{barry1999tactical,
  title={The tactical use of emotion in negotiation},
  author={Barry, Bruce},
  journal={Research on negotiation in organizations},
  volume={7},
  pages={93--124},
  year={1999},
  publisher={JAI PRESS LTD.}
}

@article{van2022moving,
  title={Moving the self and others to do good: The emotional underpinnings of prosocial behavior},
  author={Van Kleef, Gerben A and Lelieveld, Gert-Jan},
  journal={Current opinion in psychology},
  volume={44},
  pages={80--88},
  year={2022},
  publisher={Elsevier}
}

@article{lawler1999bringing,
  title={Bringing emotions into social exchange theory},
  author={Lawler, Edward J and Thye, Shane R},
  journal={Annual review of sociology},
  volume={25},
  number={1},
  pages={217--244},
  year={1999},
  publisher={Annual Reviews 4139 El Camino Way, PO Box 10139, Palo Alto, CA 94303-0139, USA}
}

@article{yurtsever2004emotional,
  title={Emotional regulation strategies and negotiation},
  author={Yurtsever, G{\"u}l{\c{c}}imen},
  journal={Psychological Reports},
  volume={95},
  number={3},
  pages={780--786},
  year={2004},
  publisher={SAGE Publications Sage CA: Los Angeles, CA}
}

@inproceedings{wang2023self,
  title={Self-Instruct: Aligning Language Models with Self-Generated Instructions},
  author={Wang, Yizhong and Kordi, Yeganeh and Mishra, Swaroop and Liu, Alisa and Smith, Noah A and Khashabi, Daniel and Hajishirzi, Hannaneh},
  booktitle={Proceedings of the 61st Annual Meeting of the Association for Computational Linguistics (Volume 1: Long Papers)},
  pages={13484--13508},
  year={2023}
}

@article{chawla2021casino,
  title={CaSiNo: A corpus of campsite negotiation dialogues for automatic negotiation systems},
  author={Chawla, Kushal and Ramirez, Jaysa and Clever, Rene and Lucas, Gale and May, Jonathan and Gratch, Jonathan},
  journal={arXiv preprint arXiv:2103.15721},
  year={2021}
}

@inproceedings{konovalov2016negochat,
  title={The negochat corpus of human-agent negotiation dialogues},
  author={Konovalov, Vasily and Artstein, Ron and Melamud, Oren and Dagan, Ido},
  booktitle={Proceedings of the Tenth International Conference on Language Resources and Evaluation (LREC'16)},
  pages={3141--3145},
  year={2016}
}

@article{mchugh2012interrater,
  title={Interrater reliability: the kappa statistic},
  author={McHugh, Mary L},
  journal={Biochemia medica},
  volume={22},
  number={3},
  pages={276--282},
  year={2012},
  publisher={Medicinska naklada}
}

@article{van2008emotion,
  title={Emotion in conflict and negotiation: Introducing the emotions as social information (EASI) model},
  author={Van Kleef, Gerben A and others},
  journal={Research companion to emotion in organizations},
  pages={392--404},
  year={2008}
}

@article{kelley1984iterative,
  title={An iterative design methodology for user-friendly natural language office information applications},
  author={Kelley, John F},
  journal={ACM Transactions on Information Systems (TOIS)},
  volume={2},
  number={1},
  pages={26--41},
  year={1984},
  publisher={ACM New York, NY, USA}
}

@article{mayer2003measuring,
  title={Measuring emotional intelligence with the MSCEIT V2. 0.},
  author={Mayer, John D and Salovey, Peter and Caruso, David R and Sitarenios, Gill},
  journal={Emotion},
  volume={3},
  number={1},
  pages={97},
  year={2003},
  publisher={American Psychological Association}
}

@article{boothby2023embracing,
  title={Embracing complexity: A review of negotiation research},
  author={Boothby, Erica J and Cooney, Gus and Schweitzer, Maurice E},
  journal={Annual Review of Psychology},
  volume={74},
  number={2023},
  pages={299--332},
  year={2023},
  publisher={Annual Reviews}
}

@article{rafailov2023direct,
  title={Direct preference optimization: Your language model is secretly a reward model},
  author={Rafailov, Rafael and Sharma, Archit and Mitchell, Eric and Manning, Christopher D and Ermon, Stefano and Finn, Chelsea},
  journal={Advances in neural information processing systems},
  volume={36},
  pages={53728--53741},
  year={2023}
}

@book{rubin2013social,
  title={The social psychology of bargaining and negotiation},
  author={Rubin, Jeffrey Z and Brown, Bert R},
  year={2013},
  publisher={Elsevier}
}

@article{sebenius1992negotiation,
  title={Negotiation analysis: A characterization and review},
  author={Sebenius, James K},
  journal={Management science},
  volume={38},
  number={1},
  pages={18--38},
  year={1992},
  publisher={INFORMS}
}

@article{bazerman2000negotiation,
  title={Negotiation},
  author={Bazerman, Max H and Curhan, Jared R and Moore, Don A and Valley, Kathleen L},
  journal={Annual review of psychology},
  volume={51},
  number={1},
  pages={279--314},
  year={2000},
  publisher={Annual Reviews 4139 El Camino Way, PO Box 10139, Palo Alto, CA 94303-0139, USA}
}

@article{spector1977negotiation,
  title={Negotiation as a psychological process},
  author={Spector, Bertram I},
  journal={Journal of Conflict Resolution},
  volume={21},
  number={4},
  pages={607--618},
  year={1977},
  publisher={Sage Publications Sage CA: Thousand Oaks, CA}
}

@article{lawler1995structural,
  title={Structural power and emotional processes in negotiation: A social exchange approach},
  author={Lawler, Edward J and Yoon, Jeongkoo},
  year={1995}
}

@article{forgas1998feeling,
  title={On feeling good and getting your way: Mood effects on negotiator cognition and bargaining strategies.},
  author={Forgas, Joseph P},
  journal={Journal of personality and social psychology},
  volume={74},
  number={3},
  pages={565},
  year={1998},
  publisher={American Psychological Association}
}

@book{goleman2005emotional,
  title={Emotional intelligence: Why it can matter more than IQ},
  author={Goleman, Daniel},
  year={2005},
  publisher={Bantam}
}

@article{li2024enhancing,
  title={Enhancing emotional generation capability of large language models via emotional chain-of-thought},
  author={Li, Zaijing and Chen, Gongwei and Shao, Rui and Xie, Yuquan and Jiang, Dongmei and Nie, Liqiang},
  journal={arXiv preprint arXiv:2401.06836},
  year={2024}
}

@article{kroczek2025influence,
  title={The influence of persona and conversational task on social interactions with a LLM-controlled embodied conversational agent},
  author={Kroczek, Leon OH and May, Alexander and Hettenkofer, Selina and Ruider, Andreas and Ludwig, Bernd and M{\"u}hlberger, Andreas},
  journal={Computers in Human Behavior},
  pages={108759},
  year={2025},
  publisher={Elsevier}
}

@article{huang2024personality,
  title={How personality traits influence negotiation outcomes? a simulation based on large language models},
  author={Huang, Yin Jou and Hadfi, Rafik},
  journal={arXiv preprint arXiv:2407.11549},
  year={2024}
}

@inproceedings{bosse2013learning,
  title={Learning emotion regulation strategies: A cognitive agent model},
  author={Bosse, Tibor and Gerritsen, Charlotte and de Man, Jeroen and Treur, Jan},
  booktitle={2013 IEEE/WIC/ACM International Joint Conferences on Web Intelligence (WI) and Intelligent Agent Technologies (IAT)},
  volume={2},
  pages={245--252},
  year={2013},
  organization={IEEE}
}

@article{dang2024explainable,
  title={Explainable and interpretable multimodal large language models: A comprehensive survey},
  author={Dang, Yunkai and Huang, Kaichen and Huo, Jiahao and Yan, Yibo and Huang, Sirui and Liu, Dongrui and Gao, Mengxi and Zhang, Jie and Qian, Chen and Wang, Kun and others},
  journal={arXiv preprint arXiv:2412.02104},
  year={2024}
}

@article{wei2022chain,
  title={Chain-of-thought prompting elicits reasoning in large language models},
  author={Wei, Jason and Wang, Xuezhi and Schuurmans, Dale and Bosma, Maarten and Xia, Fei and Chi, Ed and Le, Quoc V and Zhou, Denny and others},
  journal={Advances in neural information processing systems},
  volume={35},
  pages={24824--24837},
  year={2022}
}

@misc{openai2024chatgpt,
  author = {OpenAI},
  title = {{ChatGPT}},
  howpublished = {\url{https://chatgpt.com/}},
  year = {2024}
}

@article{zelikman2022star,
  title={Star: Bootstrapping reasoning with reasoning},
  author={Zelikman, Eric and Wu, Yuhuai and Mu, Jesse and Goodman, Noah},
  journal={Advances in Neural Information Processing Systems},
  volume={35},
  pages={15476--15488},
  year={2022}
}

@article{gulcehre2023reinforced,
  title={Reinforced self-training (rest) for language modeling},
  author={Gulcehre, Caglar and Paine, Tom Le and Srinivasan, Srivatsan and Konyushkova, Ksenia and Weerts, Lotte and Sharma, Abhishek and Siddhant, Aditya and Ahern, Alex and Wang, Miaosen and Gu, Chenjie and others},
  journal={arXiv preprint arXiv:2308.08998},
  year={2023}
}

@article{deng2024flow,
  title={Flow-dpo: Improving llm mathematical reasoning through online multi-agent learning},
  author={Deng, Yihe and Mineiro, Paul},
  journal={arXiv preprint arXiv:2410.22304},
  year={2024}
}

@article{welch1947generalization,
  title={The generalization of ‘STUDENT'S’problem when several different population varlances are involved},
  author={Welch, Bernard L},
  journal={Biometrika},
  volume={34},
  number={1-2},
  pages={28--35},
  year={1947},
  publisher={Oxford University Press}
}

@article{singh2023beyond,
  title={Beyond human data: Scaling self-training for problem-solving with language models},
  author={Singh, Avi and Co-Reyes, John D and Agarwal, Rishabh and Anand, Ankesh and Patil, Piyush and Garcia, Xavier and Liu, Peter J and Harrison, James and Lee, Jaehoon and Xu, Kelvin and others},
  journal={arXiv preprint arXiv:2312.06585},
  year={2023}
}

@article{touvron2023llama,
  title={Llama: Open and efficient foundation language models},
  author={Touvron, Hugo and Lavril, Thibaut and Izacard, Gautier and Martinet, Xavier and Lachaux, Marie-Anne and Lacroix, Timoth{\'e}e and Rozi{\`e}re, Baptiste and Goyal, Naman and Hambro, Eric and Azhar, Faisal and others},
  journal={arXiv preprint arXiv:2302.13971},
  year={2023}
}

@inproceedings{deng2023prompting,
  title={Prompting and Evaluating Large Language Models for Proactive Dialogues: Clarification, Target-guided, and Non-collaboration},
  author={Deng, Yang and Liao, Lizi and Chen, Liang and Wang, Hongru and Lei, Wenqiang and Chua, Tat-Seng},
  booktitle={Findings of the Association for Computational Linguistics: EMNLP 2023},
  pages={10602--10621},
  year={2023}
}

@misc{schulman2017proximalpolicyoptimizationalgorithms,
      title={Proximal Policy Optimization Algorithms}, 
      author={John Schulman and Filip Wolski and Prafulla Dhariwal and Alec Radford and Oleg Klimov},
      year={2017},
      eprint={1707.06347},
      archivePrefix={arXiv},
      primaryClass={cs.LG},
      url={https://arxiv.org/abs/1707.06347}, 
}

@article{deng2023plug,
  title={Plug-and-play policy planner for large language model powered dialogue agents},
  author={Deng, Yang and Zhang, Wenxuan and Lam, Wai and Ng, See-Kiong and Chua, Tat-Seng},
  journal={arXiv preprint arXiv:2311.00262},
  year={2023}
}

@article{fu2023improving,
  title={Improving language model negotiation with self-play and in-context learning from ai feedback},
  author={Fu, Yao and Peng, Hao and Khot, Tushar and Lapata, Mirella},
  journal={arXiv preprint arXiv:2305.10142},
  year={2023}
}

@article{zhang2023ask,
  title={Ask an expert: Leveraging language models to improve strategic reasoning in goal-oriented dialogue models},
  author={Zhang, Qiang and Naradowsky, Jason and Miyao, Yusuke},
  journal={arXiv preprint arXiv:2305.17878},
  year={2023}
}

@inproceedings{deng2024plug,
  title={Plug-and-Play Policy Planner for Large Language Model Powered Dialogue Agents},
  author={Deng, Yang and Zhang, Wenxuan and Lam, Wai and Ng, See-Kiong and Chua, Tat-Seng},
  booktitle={ICLR},
  year={2024}
}

@article{ouyang2022training,
  title={Training language models to follow instructions with human feedback},
  author={Ouyang, Long and Wu, Jeffrey and Jiang, Xu and Almeida, Diogo and Wainwright, Carroll and Mishkin, Pamela and Zhang, Chong and Agarwal, Sandhini and Slama, Katarina and Ray, Alex and others},
  journal={Advances in neural information processing systems},
  volume={35},
  pages={27730--27744},
  year={2022}
}

@article{zhang2019bertscore,
  title={Bertscore: Evaluating text generation with bert},
  author={Zhang, Tianyi and Kishore, Varsha and Wu, Felix and Weinberger, Kilian Q and Artzi, Yoav},
  journal={arXiv preprint arXiv:1904.09675},
  year={2019}
}

@article{li2015diversity,
  title={A diversity-promoting objective function for neural conversation models},
  author={Li, Jiwei and Galley, Michel and Brockett, Chris and Gao, Jianfeng and Dolan, Bill},
  journal={arXiv preprint arXiv:1510.03055},
  year={2015}
}

@article{brown1992estimate,
  title={An estimate of an upper bound for the entropy of English},
  author={Brown, Peter F and Della Pietra, Stephen A and Della Pietra, Vincent J and Lai, Jennifer C and Mercer, Robert L},
  journal={Computational Linguistics},
  volume={18},
  number={1},
  pages={31--40},
  year={1992}
}

@inproceedings{papineni2002bleu,
  title={Bleu: a method for automatic evaluation of machine translation},
  author={Papineni, Kishore and Roukos, Salim and Ward, Todd and Zhu, Wei-Jing},
  booktitle={Proceedings of the 40th annual meeting of the Association for Computational Linguistics},
  pages={311--318},
  year={2002}
}

@inproceedings{ahmad2023ina,
  title={INA: An Integrative Approach for Enhancing Negotiation Strategies with Reward-Based Dialogue Agent},
  author={Ahmad, Zishan and Saurabh, Suman and Menon, Vaishakh and Ekbal, Asif and Ramnani, Roshni and Maitra, Anutosh},
  booktitle={Findings of the Association for Computational Linguistics: EMNLP 2023},
  pages={2536--2549},
  year={2023}
}

@article{wolf2019huggingface,
  title={Huggingface's transformers: State-of-the-art natural language processing},
  author={Wolf, Thomas and Debut, Lysandre and Sanh, Victor and Chaumond, Julien and Delangue, Clement and Moi, Anthony and Cistac, Pierric and Rault, Tim and Louf, R{\'e}mi and Funtowicz, Morgan and others},
  journal={arXiv preprint arXiv:1910.03771},
  year={2019}
}

@article{kingma2014adam,
  title={Adam: A method for stochastic optimization},
  author={Kingma, Diederik P and Ba, Jimmy},
  journal={arXiv preprint arXiv:1412.6980},
  year={2014}
}

@article{holtzman2019curious,
  title={The curious case of neural text degeneration},
  author={Holtzman, Ari and Buys, Jan and Du, Li and Forbes, Maxwell and Choi, Yejin},
  journal={arXiv preprint arXiv:1904.09751},
  year={2019}
}

@article{carnevale1992negotiation,
  title={Negotiation and mediation},
  author={Carnevale, Peter J and Pruitt, Dean G and others},
  journal={Annual review of psychology},
  volume={43},
  number={1},
  pages={531--582},
  year={1992}
}

@inproceedings{yamaguchi2021dialogue,
  title={Dialogue act-based breakdown detection in negotiation dialogues},
  author={Yamaguchi, Atsuki and Iwasa, Kosui and Fujita, Katsuhide},
  booktitle={Proceedings of the 16th Conference of the European Chapter of the Association for Computational Linguistics: Main Volume},
  pages={745--757},
  year={2021}
}

@inproceedings{chawla2022opponent,
  title={Opponent Modeling in Negotiation Dialogues by Related Data Adaptation},
  author={Chawla, Kushal and Lucas, Gale and May, Jonathan and Gratch, Jonathan},
  booktitle={Findings of the Association for Computational Linguistics: NAACL 2022},
  pages={661--674},
  year={2022}
}

@article{zhao2019rethinking,
  title={Rethinking action spaces for reinforcement learning in end-to-end dialog agents with latent variable models},
  author={Zhao, Tiancheng and Xie, Kaige and Eskenazi, Maxine},
  journal={arXiv preprint arXiv:1902.08858},
  year={2019}
}

@inproceedings{zhou2019dynamic,
  title={A Dynamic Strategy Coach for Effective Negotiation},
  author={Zhou, Yiheng and He, He and Black, Alan W and Tsvetkov, Yulia},
  booktitle={Proceedings of the 20th Annual SIGdial Meeting on Discourse and Dialogue},
  pages={367--378},
  year={2019}
}

@inproceedings{he2018decoupling,
  title={Decoupling Strategy and Generation in Negotiation Dialogues},
  author={He, He and Chen, Derek and Balakrishnan, Anusha and Liang, Percy},
  booktitle={Proceedings of the 2018 Conference on Empirical Methods in Natural Language Processing},
  pages={2333--2343},
  year={2018}
}

@inproceedings{zhang2020learning,
  title={Learning Goal-oriented Dialogue Policy with opposite Agent Awareness},
  author={Zhang, Zheng and Liao, Lizi and Zhu, Xiaoyan and Chua, Tat-Seng and Liu, Zitao and Huang, Yan and Huang, Minlie},
  booktitle={Proceedings of the 1st Conference of the Asia-Pacific Chapter of the Association for Computational Linguistics and the 10th International Joint Conference on Natural Language Processing},
  pages={122--132},
  year={2020}
}

@article{yang2020improving,
  title={Improving dialog systems for negotiation with personality modeling},
  author={Yang, Runzhe and Chen, Jingxiao and Narasimhan, Karthik},
  journal={arXiv preprint arXiv:2010.09954},
  year={2020}
}

@article{priya2025we,
  title={We Argue to Agree: Towards Personality-Driven Argumentation-Based Negotiation Dialogue Systems for Tourism},
  author={Priya, Priyanshu and Dudhate, Saurav and Yasheshbhai, Desai Vishesh and Ekbal, Asif},
  journal={arXiv preprint arXiv:2509.11118},
  year={2025}
}

@inproceedings{priya2024trip,
  title={TRIP NEGOTIATOR: A Travel Persona-aware Reinforced Dialogue Generation Model for Personalized Integrative Negotiation in Tourism},
  author={Priya, Priyanshu and Yasheshbhai, Desai Vishesh and Joshi, Ratnesh Kumar and Ramnani, Roshni and Maitra, Anutosh and Sengupta, Shubhashis and Ekbal, Asif},
  booktitle={Findings of the Association for Computational Linguistics: EMNLP 2024},
  pages={16566--16595},
  year={2024}
}

@inproceedings{zhao2018sogo,
  title={SOGO: a social intelligent negotiation dialogue system},
  author={Zhao, Ran and Romero, Oscar J and Rudnicky, Alex},
  booktitle={Proceedings of the 18th International Conference on intelligent virtual agents},
  pages={239--246},
  year={2018}
}

@inproceedings{priya2025genteel,
  title={GENTEEL-NEGOTIATOR: LLM-Enhanced Mixture-of-Expert-Based Reinforcement Learning Approach for Polite Negotiation Dialogue},
  author={Priya, Priyanshu and Chigrupaatii, Rishikant and Firdaus, Mauajama and Ekbal, Asif},
  booktitle={Proceedings of the AAAI Conference on Artificial Intelligence},
  volume={39},
  number={23},
  pages={25010--25018},
  year={2025}
}

@article{morris2000emotions,
  title={How emotions work: The social functions of emotional expression in negotiations},
  author={Morris, Michael W and Keltner, Dacher},
  journal={Research in organizational behavior},
  volume={22},
  pages={1--50},
  year={2000},
  publisher={Elsevier}
}

@article{douglas2014attitude,
  title={Attitude and response to emotion in dispute resolution: The experience of mediators},
  author={Douglas, Kathy and Coburn, Clare},
  journal={Flinders LJ},
  volume={16},
  pages={111},
  year={2014},
  publisher={HeinOnline}
}

@inproceedings{chawla2021towards,
  title={Towards emotion-aware agents for negotiation dialogues},
  author={Chawla, Kushal and Clever, Rene and Ramirez, Jaysa and Lucas, Gale and Gratch, Jonathan},
  booktitle={2021 9th International Conference on Affective Computing and Intelligent Interaction (ACII)},
  pages={1--8},
  year={2021},
  organization={IEEE}
}

@article{chawla2023towards,
  title={Towards emotion-aware agents for improved user satisfaction and partner perception in negotiation dialogues},
  author={Chawla, Kushal and Clever, Rene and Ramirez, Jaysa and Lucas, Gale M and Gratch, Jonathan},
  journal={IEEE Transactions on Affective Computing},
  volume={15},
  number={2},
  pages={433--444},
  year={2023},
  publisher={IEEE}
}

@inproceedings{keskin2024negotiator,
  title={NEGOTIATOR: a comprehensive framework for human-agent negotiation integrating preferences, interaction, and emotion},
  author={Keskin, Mehmet Onur and Buzcu, Berk and Ko{\c{c}}yi{\u{g}}it, Berkecan and {\c{C}}akan, Umut and Do{\u{g}}ru, An{\i}l and Aydo{\u{g}}an, Reyhan},
  booktitle={Proceedings of the Thirty-Third International Joint Conference on Artificial Intelligence (IJCAI-24)},
  pages={8700--8703},
  year={2024}
}

@article{lewis2017deal,
  title={Deal or no deal? end-to-end learning for negotiation dialogues},
  author={Lewis, Mike and Yarats, Denis and Dauphin, Yann N and Parikh, Devi and Batra, Dhruv},
  journal={arXiv preprint arXiv:1706.05125},
  year={2017}
}

@article{minaee2024large,
  title={Large language models: A survey},
  author={Minaee, Shervin and Mikolov, Tomas and Nikzad, Narjes and Chenaghlu, Meysam and Socher, Richard and Amatriain, Xavier and Gao, Jianfeng},
  journal={arXiv preprint arXiv:2402.06196},
  year={2024}
}

@article{gandhi2023strategic,
  title={Strategic reasoning with language models},
  author={Gandhi, Kanishk and Sadigh, Dorsa and Goodman, Noah D},
  journal={arXiv preprint arXiv:2305.19165},
  year={2023}
}

@article{abdelnabi2023llm,
  title={Llm-deliberation: Evaluating llms with interactive multi-agent negotiation games},
  author={Abdelnabi, Sahar and Gomaa, Amr and Sivaprasad, Sarath and Sch{\"o}nherr, Lea and Fritz, Mario},
  journal={arXiv preprint arXiv:2309.17234},
  year={2023}
}

@article{li2023toward,
  title={Toward explainable dialogue system using two-stage response generation},
  author={Li, Shaobo and Sun, Chengjie and Xu, Zhen and Tiwari, Prayag and Liu, Bingquan and Gupta, Deepak and Shankar, K and Ji, Zhenzhou and Wang, Mingjiang},
  journal={ACM Transactions on Asian and Low-Resource Language Information Processing},
  volume={22},
  number={3},
  pages={1--18},
  year={2023},
  publisher={ACM New York, NY}
}

@article{yang2024xagents,
  title={XAgents: A Framework for Interpretable Rule-Based Multi-Agents Cooperation},
  author={Yang, Hailong and Gu, Mingxian and Zhao, Renhuo and Hu, Fuping and Deng, Zhaohong and Chen, Yitang},
  journal={arXiv preprint arXiv:2411.13932},
  year={2024}
}

@article{fuhrman2022review,
  title={A review of explainable and interpretable AI with applications in COVID-19 imaging},
  author={Fuhrman, Jordan D and Gorre, Naveena and Hu, Qiyuan and Li, Hui and El Naqa, Issam and Giger, Maryellen L},
  journal={Medical Physics},
  volume={49},
  number={1},
  pages={1--14},
  year={2022},
  publisher={Wiley Online Library}
}

@article{maier2004emotional,
  title={Emotional intelligence: theory findings, and implications. Psychological Inquriry, 15, 197-215},
  author={Maier, JD and Salovey, P and Caruso, DR},
  journal={Psychological Science},
  volume={9},
  number={5},
  pages={331--339},
  year={2004}
}

@article{feng2025emotionally,
  title={Emotionally Intelligent Task-oriented Dialogue Systems: Architecture, Representation, and Optimisation},
  author={Feng, Shutong and Lin, Hsien-chin and Lubis, Nurul and van Niekerk, Carel and Heck, Michael and Ruppik, Benjamin and Vukovic, Renato and Ga{\v{s}}i{\'c}, Milica},
  journal={arXiv preprint arXiv:2507.01594},
  year={2025}
}

@inproceedings{kasbouya2025emotional,
  title={Emotional intelligence in large language models: Fine-tuning methods, challenges, and applications},
  author={Kasbouya, Mohammed and Sael, Nawal},
  booktitle={International Conference on intelligent systems and digital applications},
  pages={348--359},
  year={2025},
  organization={Springer}
}

@article{zhang2022automatic,
  title={Automatic Chain of Thought Prompting in Large Language Models},
  author={Zhang, Zhuosheng and Zhang, Aston and Li, Mu and Smola, Alex},
  journal={arXiv preprint arXiv:2210.03493},
  year={2022}
}

@article{yang2023psycot,
  title={Psycot: Psychological questionnaire as powerful chain-of-thought for personality detection},
  author={Yang, Tao and Shi, Tianyuan and Wan, Fanqi and Quan, Xiaojun and Wang, Qifan and Wu, Bingzhe and Wu, Jiaxiang},
  journal={arXiv preprint arXiv:2310.20256},
  year={2023}
}

@article{lai2024step,
  title={Step-dpo: Step-wise preference optimization for long-chain reasoning of llms},
  author={Lai, Xin and Tian, Zhuotao and Chen, Yukang and Yang, Senqiao and Peng, Xiangru and Jia, Jiaya},
  journal={arXiv preprint arXiv:2406.18629},
  year={2024}
}

@article{bai2022training,
  title={Training a helpful and harmless assistant with reinforcement learning from human feedback},
  author={Bai, Yuntao and Jones, Andy and Ndousse, Kamal and Askell, Amanda and Chen, Anna and DasSarma, Nova and Drain, Dawn and Fort, Stanislav and Ganguli, Deep and Henighan, Tom and others},
  journal={arXiv preprint arXiv:2204.05862},
  year={2022}
}

@inproceedings{trung2024reft,
  title={Reft: Reasoning with reinforced fine-tuning},
  author={Trung, Luong and Zhang, Xinbo and Jie, Zhanming and Sun, Peng and Jin, Xiaoran and Li, Hang},
  booktitle={Proceedings of the 62nd Annual Meeting of the Association for Computational Linguistics (Volume 1: Long Papers)},
  pages={7601--7614},
  year={2024}
}

@article{wang2025reinforcement,
  title={Reinforcement learning for reasoning in large language models with one training example},
  author={Wang, Yiping and Yang, Qing and Zeng, Zhiyuan and Ren, Liliang and Liu, Liyuan and Peng, Baolin and Cheng, Hao and He, Xuehai and Wang, Kuan and Gao, Jianfeng and others},
  journal={arXiv preprint arXiv:2504.20571},
  year={2025}
}

@article{scudder1965probability,
  title={Probability of error of some adaptive pattern-recognition machines},
  author={Scudder, Henry},
  journal={IEEE Transactions on Information Theory},
  volume={11},
  number={3},
  pages={363--371},
  year={1965},
  publisher={IEEE}
}

@article{amini2025self,
  title={Self-training: A survey},
  author={Amini, Massih-Reza and Feofanov, Vasilii and Pauletto, Loic and Hadjadj, Lies and Devijver, Emilie and Maximov, Yury},
  journal={Neurocomputing},
  volume={616},
  pages={128904},
  year={2025},
  publisher={Elsevier}
}

@article{lee2024self,
  title={Self-Training Meets Consistency: Improving LLMs' Reasoning with Consistency-Driven Rationale Evaluation},
  author={Lee, Jaehyeok and Sakaguchi, Keisuke and Bak, JinYeong},
  journal={arXiv preprint arXiv:2411.06387},
  year={2024}
}

@article{wu2024rethinking,
  title={Rethinking chain-of-thought from the perspective of self-training},
  author={Wu, Zongqian and Xu, Baoduo and Cui, Ruochen and Zhan, Mengmeng and Zhu, Xiaofeng and Feng, Lei},
  journal={arXiv preprint arXiv:2412.10827},
  year={2024}
}

@misc{openai2025gpt5,
  title        = {GPT-5 System Card},
  author       = {OpenAI},
  year         = {2025},
  howpublished = {\url{https://cdn.openai.com/gpt-5-system-card.pdf}},
  note         = {Accessed: October 2025}
}
\newpage
\appendix

\section*{Appendix}
\label{sec:appendix}

\section{Dataset Details}
\label{dataset_details}

\subsection{Prompt Templates}
\label{prompt_appendix}
The prompt template used to generate the negotiation dialogue with ENS-CoT rationales is given below:

\begin{tcolorbox}[
    enhanced,
    colback=gray!5,
    colframe=gray!80,
    coltitle=black,        
    sharp corners=all,
    boxrule=0.8pt,
    title=\textbf{Prompt for Negotiation Dialogue Generation with ENS-CoT Rationales. Variables enclosed in curly brackets serve as placeholders that are populated when the template is instantiated.},
    fonttitle=\bfseries,
    breakable
]
{\small
Your task is to generate a negotiation dialogue between the user and the agent for the $\{$Negotiation Scenario$\}$. You need to generate the user's utterance and mark the expressed emotion (EM), emotion trigger (ET), individual assessment (IA), perspective shift (PS), mindset transformation (MT), strategy selection (SS), strategy reason (SR), and agent's response based on the selected strategy (RG). \\

User's Utterance: $\langle$Turn contributed by the user$\rangle$ User's utterance should begin with `\{\textit{User:} $\dots$\}'

Emotion: $\langle$EM definition$\rangle$ Emotion should begin with `\{\textit{The user feels} $\dots$\}'

Emotional Trigger: $\langle$ET definition$\rangle$ Emotional Trigger should begin with `\{\textit{User's} `Emotion' \textit{is triggered by} $\dots$\}'

Individual Assessment: $\langle$IA definition$\rangle$ Individual Assessment should begin with `\{\textit{The user thinks} $\dots$\}'

Perspective Shift: $\langle$PS definition$\rangle$ Perspective Shift should begin with `\{\textit{Enable the user to consider the situation from a different angle:} $\dots$\}'

Mindset Transformation: $\langle$MT definition$\rangle$ Mindset Transformation should begin with `\{\textit{Enable the user to think about reframing the belief:} $\dots$\}' 

Strategy Selection: $\langle$SS definition$\rangle$ Strategy Selection should begin with `\{\textit{The agent chooses} $\dots$\}'

Strategy Reason: $\langle$SR definition$\rangle$ Strategy Reason should begin with `\{\textit{To} $\dots$, \textit{the agent uses} $\dots$\}'

Response Generation: $\langle$RG definition$\rangle$ Agent's response should begin with \{\textit{Agent:} $\dots$\}'\\

The user should use the following emotions: $\langle$\textit{list of emotions}$\rangle$

The user should use the following emotion-aware negotiation strategies: $\langle$\textit{list of strategies}$\rangle$

Examples: \\

$\langle$\textit{Few-shot exemplars}$\rangle$ \\

Please adhere precisely to the format provided above, ensuring that no components are omitted.\\

--- To be generated ---\\

$\{$Generated Dialogue$\}$\\

}
\end{tcolorbox}
The prompt template employed to generate negotiation scenarios is as follows:

\begin{tcolorbox}[
    enhanced,
    colback=gray!5,
    colframe=gray!80,
    coltitle=black,        
    sharp corners=all,
    boxrule=0.8pt,
    title=\textbf{Prompt for Negotiation Scenario Generation. Variables enclosed in curly brackets serve as placeholders that are populated when the template is instantiated.},
    fonttitle=\bfseries,
    breakable
]
{\small
Your task is to generate the scenario, summarizing the negotiation setup, including the task, roles, issues, constraints, and goals underlying the provided dialogue.\\

Examples: (Optional) \\

$\langle$\textit{Few-shot exemplars}$\rangle$ \\

The dialogue is as follows:\\

$\langle$\textit{Seed Negotiation Dialogue}$\rangle$ \\

--- To be generated ---\\

$\{$Generated Negotiation Scenario$\}$\\

}
\end{tcolorbox}

\subsection{Examples for Emotion-aware Negotiation Strategies}
Table \ref{strategy_examples} provides the example utterances for different emotion-aware negotiation strategies from the \texttt{JobNego} dataset.

\subsection{Guidelines for Designing Sample ENS-CoT-enhanced Dialogues}
The annotators are instructed to adhere to the following guidelines during dialogue creation:

\begin{itemize}[leftmargin=1.2em,itemsep=0pt]
    \item To enhance understanding of negotiation and ensure the application of best negotiation practices, we follow \cite{chawla2021casino} and require the annotators to watch a tutorial on principled negotiation strategies before beginning their tasks;
    \item Begin negotiations with offers that leave room for concessions and adjustments, allowing both sides to explore trade-offs;
    \item Actively elicit and discuss the other party’s preferences, priorities, and constraints to identify opportunities for integrative outcomes;
    \item Clearly communicate requirements and rationales, grounding arguments in realistic job interview (e.g., salary, promotion, benefits) or resource allocation (e.g., distribution of supplies, task responsibilities) scenarios;
    \item Recognize and respond to the emotional states expressed by the user, ensuring that the agent applies ENS-CoT reasoning to \textit{perceive}, \textit{understand}, \textit{use}, and \textit{manage} emotions in a transparent manner;
    \item Ensure that each agent response is paired with a concise ENS-CoT annotation, explicitly justifying how a particular emotion-aware negotiation strategy is chosen;
    \item Aim for outcomes that balance self-interest with fairness, recognizing that perceptions of equity and rationality may differ across negotiating parties;
    \item Use professional, and empathetic language consistently, especially in situations of disagreement or emotional tension, to maintain rapport and trust.
\end{itemize}

\subsection{Details of Dataset Quality Evaluation Criteria}
\begin{enumerate}[leftmargin=1.2em,itemsep=0pt, parsep=0pt, topsep=0pt, partopsep=0pt]
\setlength{\itemsep}{0pt}
\setlength{\parskip}{0pt}
\setlength{\parsep}{0pt}
    \item \textbf{Emotional Intelligence (EI)}: Is the response appropriate to the user's expressed emotion, reducing misattunement or escalation?
    \item \textbf{Strategy Appropriateness (SA)}: Does the agent's utterance operationalize the emotion-aware negotiation strategy?
    \item \textbf{Interpretability (IN)}: Is the ENS-CoT rationale well-formed, complete, and useful?
    \item \textbf{Fairness (F)}: Does the final outcome reflect a balanced, and win-win situation?
    \item \textbf{Coherence (C)}: Is the overall dialogue well-structured and logical?
    \item \textbf{Naturalness (N)}: Does the dialogue resembles the human negotiation?
    \item \textbf{Interestingness (I)}: Is the dialogue engaging and rich in content to retain the user's interest throughout the negotiation? 
\end{enumerate}

\begin{table*}[t]\footnotesize
\centering
\small
\begin{tabular}{p{3.2cm}|p{13cm}}
\hline
\textbf{Emotion-aware Negotiation Strategy} & \textbf{Examples} \\
\hline
\textbf{Savoring} & \textit{It's great to see your excitement, and we appreciate your clear expectations. Let's discuss the salary range for the project manager role. } \\ \hline
\textbf{Positive Reinforcement} & \textit{Great, thank you for being open to compromise. It's important for us to work together to ensure a productive and motivating work environment. Let's continue discussing your needs and expectations for the company car and workday} \\ \hline
\textbf{Expressing Optimism} & \textit{We appreciate your confidence in your abilities and your willingness to work with us. Let's discuss the promotion track and how we can align our expectations for mutual success.} \\ \hline
\textbf{Cognitive Reappraisal} & \textit{Thank you for sharing your concerns. Let's explore other opportunities that match your skills and experience.}  \\ \hline
\textbf{Positive Framing} & \textit{I understand your concerns about the workday length. Let's explore how a shorter workday could lead to improved productivity and work-life balance while ensuring your career growth expectations are met.} \\ \hline
\textbf{Emotion Diffusion} & \textit{I am glad we could reach an agreement. I look forward to working with you and discussing opportunities for professional growth.} \\ \hline
\textbf{Expressive Suppression} & \textit{I appreciate your honesty. Let's discuss the employer's offer and how it can align with your career goals. How do you see yourself growing within the company?}\\ \hline
\textbf{Active Listening} & \textit{Thank you for sharing your expectations. I understand your desire for a company car. Let's discuss what we can do to ensure a mutually beneficial agreement.} \\ \hline
\textbf{Perspective-Taking} & \textit{Thank you for sharing your thoughts on the position. Let's explore the terms of the offer further to find a mutually beneficial solution.  Can we discuss the salary range for the project manager role, and how it aligns with your expectations?} \\ \hline
\textbf{Problem Solving} & \textit{I understand your concerns about work hours. How about we discuss possible compromises that can work for both of us?} \\ \hline
\textbf{Escalate Assurance} & \textit{I understand your concerns, but we are willing to discuss a salary range of 80-100,000. This is an opportunity for us to align our expectations and find a mutually beneficial agreement.} \\ \hline
\textbf{No Strategy} & \textit{Thank you for sharing your expectations. However, we cannot accommodate your requested salary at this time. We appreciate your understanding and would be willing to discuss alternative options.} \\ \hline

\end{tabular}
\caption{The example utterances of different strategies in the \texttt{JobNego} dataset.}
\label{strategy_examples}
\end{table*}

\subsection{Dataset Distribution and Topic Analysis}
\label{joint_topic}
To assess the balance, diversity, and thematic coverage of the proposed datasets, we conduct a comprehensive analysis of (i) the joint distribution of emotions and negotiation strategies, and (ii) scenario-level topical diversity for both JobNego and ResNego.

\subsubsection{Emotion-Strategy Distribution}
\begin{figure}[t]
    \centering
    \begin{subfigure}[t]{0.48\linewidth}
        \centering
        \includegraphics[width=\linewidth]{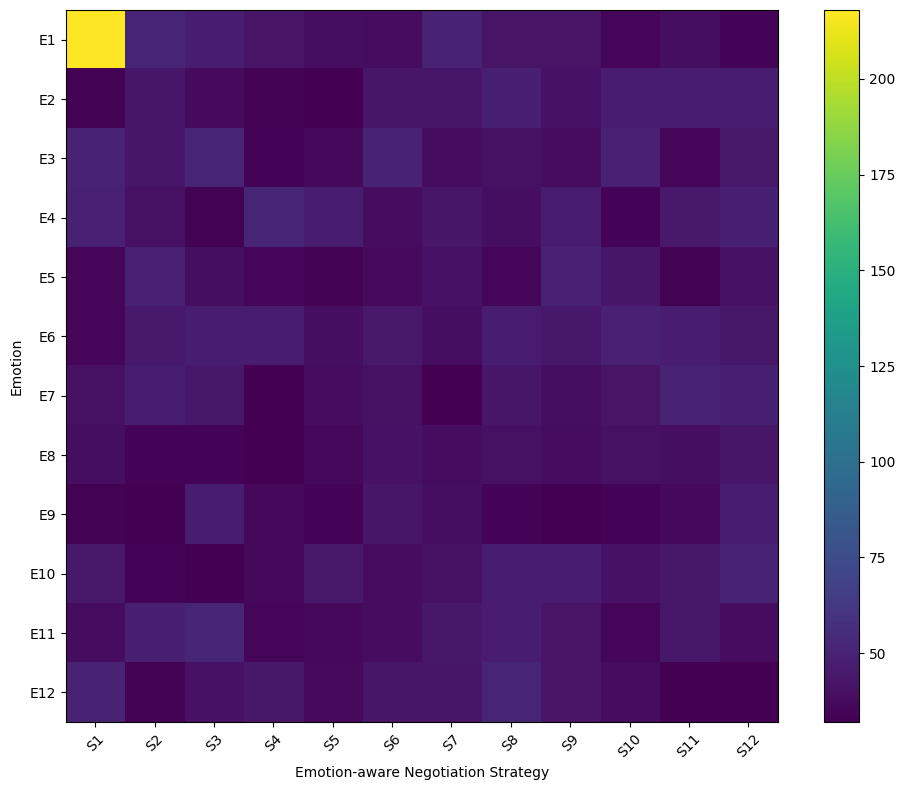}
        \caption{JobNego}
        \label{fig:jobnego_heatmap}
    \end{subfigure}
    \hfill
    \begin{subfigure}[t]{0.48\linewidth}
        \centering
        \includegraphics[width=\linewidth]{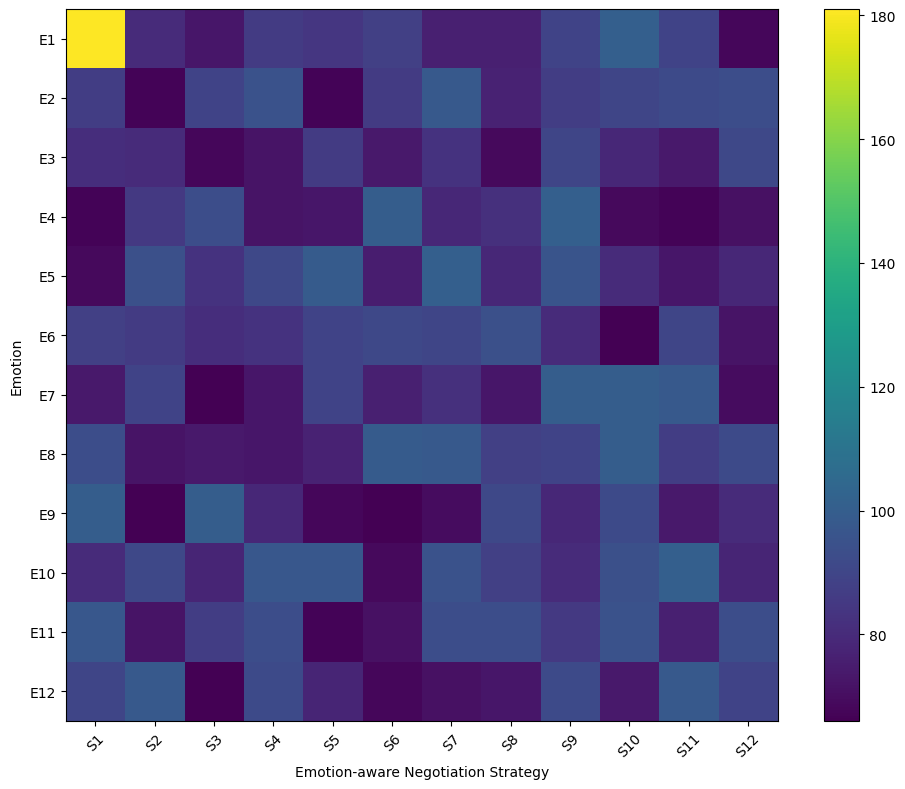}
        \caption{ResNego}
        \label{fig:resnego_heatmap}
    \end{subfigure}

    \caption{Distribution of emotion-strategy pairs in the JobNego and ResNego datasets. Emotions E1--E12 correspond to \textit{joy, confidence, positivity, gratitude, trust, surprise, anger, disappointment, frustration, fear, anxiety}, and \textit{neutral}, while emotion-aware negotiation strategies S1--S12 correspond to \textit{savoring, positive reinforcement, expressing optimism, cognitive reappraisal, positive framing, emotion diffusion, expressive suppression, active listening, perspective-taking, problem solving, escalate assurance}, and \textit{no strategy}.}
    \label{fig:emotion_strategy_heatmaps}
\end{figure}

Our datasets are constructed using 12 emotion categories and 12 emotion-aware negotiation strategies, yielding 144 possible emotion–strategy combinations. To examine distributional balance, we analyze the frequency of all combinations across both datasets. Figure~\ref{fig:emotion_strategy_heatmaps} presents a heatmap visualization of the joint emotion-strategy distribution for JobNego and ResNego. 

All emotion–strategy pairs appear in both datasets, and their frequencies are broadly distributed, with no single combination disproportionately dominating the corpus. This balanced coverage arises from ENS-CoT's context-dependent reasoning, where emotion perception and strategy selection are dynamically conditioned on the negotiation scenario rather than sampled uniformly. The observed distribution confirms that the datasets are not biased toward particular emotional states or negotiation strategies and instead capture a wide range of emotionally grounded negotiation behaviors.

\subsubsection{Topic Diversity and Thematic Coverage}
To evaluate thematic comprehensiveness, we perform a scenario-level topic analysis based on negotiation objectives, discussed resources, and contractual elements. This analysis confirms that both datasets span a diverse range of realistic negotiation contexts within their respective domains. The topic-wise distribution of dialogues for both datasets is summarized in Table~\ref{tab:topic_distribution}.

\paragraph{ResNego.}
ResNego captures multifaceted resource-allocation negotiations characteristic of survival, camping, and shared-resource environments. As shown in Table~\ref{tab:topic_distribution}, the dataset covers:
\begin{itemize}[leftmargin=1.2em,itemsep=0pt, parsep=0pt, topsep=0pt, partopsep=0pt]
    \item \textbf{Resource Requests \& Immediate Needs:} food, water, shelter materials, and essential supplies;
    \item \textbf{Firewood \& Energy Resources:} heating, cooking, and energy sourcing;
    \item \textbf{Water Access \& Safety Conditions:} purification, drought-related stressors, and drinking safety;
    \item \textbf{Group, Family \& Care Responsibilities:} negotiations involving children, seniors, dependents, and group-level prioritization;
    \item \textbf{Animals \& Campsite Safety:} wildlife risks, food storage concerns, and nighttime safety.
\end{itemize}

\paragraph{JobNego.}
JobNego reflects realistic professional negotiation scenarios commonly encountered in job interviews and workplace discussions. As summarized in Table~\ref{tab:topic_distribution}, the dataset includes:
\begin{itemize}[leftmargin=1.2em,itemsep=0pt, parsep=0pt, topsep=0pt, partopsep=0pt]
    \item \textbf{Salary \& Compensation:} base pay, compensation adjustments, and salary expectations;
    \item \textbf{Candidate Qualifications \& Fit:} experience, skills, role suitability, and team alignment;
    \item \textbf{Negotiation \& Agreement Process:} contract drafting, term confirmation, and agreement refinement;
    \item \textbf{Career Growth \& Future Prospects:} promotion pathways, advancement expectations, and long-term career planning.
\end{itemize}

Overall, these topic distributions demonstrate that both JobNego and ResNego are thematically comprehensive, covering a broad and realistic spectrum of negotiation needs. Combined with balanced emotion--strategy coverage and stable dialogue lengths, this analysis confirms that the datasets are representative and well-suited for studying emotionally intelligent and interpretable negotiation behavior.

\begin{table}[t]
\centering
\small
\begin{tabular}{l c}
\hline
\textbf{Topic} & \textbf{\# Dialogues} \\
\hline
\multicolumn{2}{c}{\textbf{JobNego}} \\
\hline
Salary \& Compensation & 370 \\
Candidate Qualifications \& Fit & 235 \\
Negotiation \& Agreement Process & 115 \\
Career Growth \& Future Prospects & 120 \\
\hline
\textbf{Total} & 840 \\
\hline
\multicolumn{2}{c}{\textbf{ResNego}} \\
\hline
Resource Requests \& Immediate Needs & 495 \\
Firewood \& Energy Resources & 297 \\
Water Access \& Safety Conditions & 330 \\
Group, Family \& Care Responsibilities & 280 \\
Animals \& Campsite Safety & 246 \\
\hline
\textbf{Total} & 1{,}648 \\
\hline
\end{tabular}
\caption{Topic-wise distribution of dialogues across the JobNego and ResNego datasets.}
\label{tab:topic_distribution}
\end{table}

\section{Experiment Details}
\label{appendix_experiment}
\subsection{Implementation Details} 
All the experiments are implemented in \href{https://pytorch.org/}{PyTorch} using transformer architectures from the Hugging Face library \cite{wolf2019huggingface}, and trained on an NVIDIA A100 80GB GPU (CUDA 12.8). During dataset preparation, ChatGPT model, \texttt{GPT-4o-mini} is used via API for generating scenarios and dialogues with Top-\textit{p} sampling ($p=0.95$) \cite{holtzman2019curious} and temperature set to $0.9$. During supervised initialization, the model is fine-tuned for 2 epochs with batch size 2. In Step 2 (a), 5 rationales per context are generated with temperature 0.7, and $\beta=0.1$ in $\mathcal{L}_{\text{PL}}(\theta')$. In Step 2 (b), 3 rationales per context are sampled from the DPO-refined model $\pi_{{\theta}'}$ with temperature 0.7. Models are optimized using AdamW \citep{kingma2014adam} with weight decay 0.01, gradient clipping 1.0, and a cosine LR scheduler at $3e^{-7}$ with 0.1 warm-up ratio. During decoding, the temperature is set to 0.7. For Step 2, the number of iterations is empirically set to $3$, with thresholds $\tau_1=0.8$, $\tau_2=0.4$, and $\tau_3=0.8$. For OpenAI's GPT-5 experiment, we use \texttt{gpt-5-mini} via API \cite{openai2025gpt5}.

\subsubsection{Sensitivity Analysis of $\tau_1, \tau_2, \tau_3$}

We select the thresholds $\tau_1$, $\tau_2$, and $\tau_3$ empirically by analyzing the cosine similarity distribution between generated and reference responses on the development set. Specifically, we choose $\tau_1$ and $\tau_3$ as higher similarity thresholds to identify high-confidence positive samples that reliably preserve both the intended emotion and the corresponding emotion-aware negotiation strategy, while $\tau_2$ is chosen as a lower similarity threshold to capture clearly suboptimal or misaligned responses. A margin between these thresholds is intentionally maintained to exclude ambiguous pairs that may introduce noise during preference learning. Empirically, responses with cosine similarity above $0.8$ consistently maintain emotional and strategic alignment, whereas responses with similarity below $0.4$ frequently exhibit semantic drift, such as generic replies or incorrect emotion or strategy usage.

To assess sensitivity, we evaluate $\tau_1, \tau_3 \in \{0.75, 0.80, 0.85\}$ and $\tau_2 \in \{0.35, 0.40, 0.45\}$. As shown in Table \ref{tab:tau_analysis}, across both JobNego and ResNego datasets, the configuration $\tau_1 = \tau_3 = 0.8$ and $\tau_2 = 0.4$ consistently yields the best performance on EA, ENSC, and BS-F1, while neighboring settings result in only minor performance variations. These results indicate that the proposed approach is not overly sensitive to the exact threshold values, and that selecting higher values for $\tau_1$ and $\tau_3$ and a lower value for $\tau_2$ based on the similarity distribution provides a stable and reproducible criterion for preference labeling and pseudo-label filtering.

\begin{table}[t]
\centering
\begin{adjustbox}{max width = 0.9\linewidth}
\begin{tabular}{cccccc}
\hline
$\tau_1$ & $\tau_2$ & $\tau_3$ & EA $\uparrow$ & ENSC $\uparrow$ & BS-F1 $\uparrow$ \\
\hline
\multicolumn{6}{c}{\textbf{JobNego}} \\
\hline
0.75 & 0.35 & 0.75 & 0.828 & 0.809 & 0.901 \\
0.75 & 0.40 & 0.75 & 0.833 & 0.813 & 0.904 \\
0.75 & 0.45 & 0.75 & 0.826 & 0.807 & 0.899 \\
0.80 & 0.35 & 0.80 & 0.836 & 0.816 & 0.907 \\
0.80 & 0.40 & 0.80 & 0.842 & 0.823 & 0.912 \\
0.80 & 0.45 & 0.80 & 0.834 & 0.814 & 0.905 \\
0.85 & 0.35 & 0.85 & 0.831 & 0.812 & 0.903 \\
0.85 & 0.40 & 0.85 & 0.837 & 0.817 & 0.908 \\
0.85 & 0.45 & 0.85 & 0.829 & 0.810 & 0.900 \\
\hline
\multicolumn{6}{c}{\textbf{ResNego}} \\
\hline
0.75 & 0.35 & 0.75 & 0.858 & 0.846 & 0.910 \\
0.75 & 0.40 & 0.75 & 0.863 & 0.851 & 0.914 \\
0.75 & 0.45 & 0.75 & 0.856 & 0.844 & 0.908 \\
0.80 & 0.35 & 0.80 & 0.868 & 0.855 & 0.918 \\
0.80 & 0.40 & 0.80 & 0.874 & 0.861 & 0.922 \\
0.80 & 0.45 & 0.80 & 0.866 & 0.853 & 0.916 \\
0.85 & 0.35 & 0.85 & 0.861 & 0.848 & 0.912 \\
0.85 & 0.40 & 0.85 & 0.870 & 0.857 & 0.919 \\
0.85 & 0.45 & 0.85 & 0.859 & 0.847 & 0.911 \\
\hline
\end{tabular}
\end{adjustbox}
\caption{Performance across different $\tau_1$ and $\tau_2$ settings. Note that $\tau_3 = \tau_1$.}
\label{tab:tau_analysis}
\end{table}

\subsection{Baseline Details}
\begin{enumerate}[label=(\roman*), itemsep=0pt, topsep=0pt, parsep=0pt, partopsep=0pt]
\item Llama-2-7b-SFT \cite{touvron2023llama}: Fine-tuning LLama-2-7b model on negotiation dialogues in a supervised setting. 
\item Llama-3.2-3b-SFT \cite{touvron2023llama}: Fine-tuning LLama-3.2-3b-instruct model on negotiation dialogues in a supervised setting.
\item Emotion-aware SFT (Llama-2-7b): Fine-tuning the LLama-2-7b model on negotiation dialogues with emotions in a supervised setting.
\item AnE \cite{zhang2023ask}: Query the LLM (GPT-4o-mini: acting as a strategy expert) with the targeted context to reason about the agent's strategy and then prompt another LLM (GPT-5-mini) to generate the response based on the selected strategy.
\item ICL-AIF \cite{fu2023improving}: Engage another LLM (GPT-4o-mini) to provide feedback that helps the agents refine their strategy (unlike AnE, the feedback is given at the dialogue level and iterates) and finally generates the response using the chosen strategy. 
\item Standard (GPT-5-mini) \cite{deng2023prompting}: Prompting GPT-5-mini using task instruction without considering any strategy.
\item Proactive (GPT-5-mini) \cite{deng2023prompting}: Prompting GPT-5-mini to first choose the strategy for the agent's turn, then generate the response conditioned on that chosen strategy.
\item ProCoT (GPT-5-mini) \cite{deng2023prompting}: Extend Proactive by prompting GPT-5-mini to generate a chain-of-thought analysis that plans the strategy for the agent's turn, then generate the response.
\item PPDPP \cite{deng2024plug}: A plug-and-play, tunable LM policy planner trained with SFT plus RL from goal-oriented AI feedback via self-play to plan the agent's strategies.
\item RLHF \cite{ouyang2022training}: Refine a model previously fine-tuned on context–response pairs via supervised learning using Proximal Policy Optimization (PPO) \cite{schulman2017proximalpolicyoptimizationalgorithms}, leveraging the same preference pairs generated in Step 2 in an RL setup with KL regularization.
\end{enumerate}

\subsection{Evaluation Metrics Details}
\paragraph{Automatic Evaluation Metrics.} Perplexity (PPL) \cite{brown1992estimate} measures how well the model predicts the next token in a response. BLEU-4 (B-4) \cite{papineni2002bleu} assesses the overlap of 4-grams between generated and ground-truth responses. Distinct-3 (D-3) \cite{li2015diversity} evaluates lexical diversity by computing the proportion of unique trigrams in the generated responses. BERTScore-F1 (BS-F1) \cite{zhang2019bertscore}\footnote{BERTScore: \url{https://huggingface.co/spaces/evaluate-metric/bertscore}}. Emotion Appropriateness (EA) evaluates the appropriateness of users' perceived emotions, and Emotion-aware Negotiation Strategy Consistency (ENSC) assesses the consistency of generated responses with the intended strategies.

\paragraph{Human Evaluation Metrics.} Fluency (F) assesses grammatical correctness and overall language quality. Coherence (C) measures whether responses are logically consistent and contextually relevant within the dialogue. Engagingness (E) evaluates how interesting and captivating the conversation is for the user. Bargaining Efficacy (BE) gauges the system's ability to bargain or present incentives that can influence the counterpart's decisions. Outcome Fairness (OF) examines whether the negotiation results in balanced and equitable agreements. Additionally, we include EA and ENSC, which mirror their automatic counterparts but are judged by humans to provide a more reliable assessment of emotion appropriateness and emotion-aware negotiation strategy consistency.

\paragraph{Human Evaluation Process.} The human evaluation is conducted by three independent evaluators\footnote{These evaluators are separate from those who participated in the dataset preparation and are compensated in accordance with institutional guidelines.}. Two of them hold Ph.D. degrees in Linguistics, while the third has a postgraduate degree in Computer Science. All evaluators have substantial prior experience with similar tasks. Before the evaluation, the annotators are briefed about negotiation scenarios. Each annotator carries out multi-turn negotiation sessions with the system for a given scenario, and completes 25 interactions each using different response sets, resulting in a total of 75 human-evaluated negotiation dialogues. After finishing the interactions, each dialogue is rated for F, C, E, EA, ENSC, BE, and OF on the scale of 1-5 (low to high). We compute Fleiss' $\kappa$ \cite{mchugh2012interrater} for all human-rated dimensions, and obtain $\kappa$ = 0.73 (F), 0.78 (C), 0.76 (E), 0.81 (EA), 0.79 (ENSC), 0.74 (BE), and 0.72 (OF), indicating substantial agreement among evaluators.

\section{Additional Analysis}
\label{analysis_appendix}

\begin{table}[!t]
\centering
\begin{adjustbox}{max width=\linewidth}
\begin{tabular}{lccccccc}
\toprule
\textbf{\# Iterations} & \textbf{PPL} $\downarrow$ & \textbf{B-4} $\uparrow$ & \textbf{D-3} $\uparrow$ & \textbf{BS-F1} $\uparrow$ & \textbf{R-LEN} $\uparrow$ & \textbf{EA} $\uparrow$ & \textbf{ENSC} $\uparrow$ \\
\hline
\multicolumn{8}{c}{\textbf{\texttt{JobNego}}} \\
\hline
0 & 5.480 & 6.401 & 0.305 & 0.899 & 25.612 & 0.815 & 0.785 \\
1 & 5.341 & 6.731 & 0.313 & 0.904 & 26.298 & 0.828 & 0.804 \\
2 & 5.254 & 6.982 & 0.318 & 0.908 & 26.813 & 0.836 & 0.815 \\
\rowcolor{lightgray}
3 & \textbf{5.196} & \textbf{7.119} & \textbf{0.322} & \textbf{0.912} & \textbf{27.108} & \textbf{0.842} & \textbf{0.823} \\
4 & 5.271 & 6.950 & 0.317 & 0.907 & 26.547 & 0.835 & 0.812 \\
\hline
\multicolumn{8}{c}{\textbf{\texttt{ResNego}}} \\
\hline
0 & 4.710 & 7.423 & 0.327 & 0.910 & 30.518 & 0.846 & 0.828 \\
1 & 4.603 & 7.864 & 0.333 & 0.915 & 31.126 & 0.860 & 0.842 \\
2 & 4.524 & 8.051 & 0.337 & 0.919 & 31.617 & 0.868 & 0.852 \\
\rowcolor{lightgray}
3 & \textbf{4.462} & \textbf{8.231} & \textbf{0.341} & \textbf{0.922} & \textbf{32.137} & \textbf{0.874} & \textbf{0.861} \\
4 & 4.531 & 8.021 & 0.336 & 0.918 & 31.443 & 0.866 & 0.848 \\
\bottomrule
\end{tabular}
\end{adjustbox}
\caption{Analysis on the impact of iterative training.}
\label{ablation_iteration}
\end{table}
\subsection{Impact of Supervised Initialization, Self-Training and DPO}
To investigate the contribution of different training components in \textsc{prisma}, we perform an ablation study by removing supervised initialization (SI), self-training (ST), and DPO steps individually. Table~\ref{ablation_methods} presents the performance results of these ablations on both datasets. The removal of SI leads to noticeable drops in B-4, D-3, and BS-F1 across both datasets, suggesting that SI is important for stable model training. The exclusion of the ST reduces both diversity and goal attainment metrics scores, showing that ST helps improve response variability, emotion appropriateness, and negotiation effectiveness. The omission of the DPO step results in further decline in performance across all the metrics on both datasets, indicating that DPO fine-tuning further refines responses to align with the intended objective of enhancing emotional intelligence in negotiation dialogues.
\begin{table}[!t]
\centering
\begin{adjustbox}{max width=\linewidth}
\begin{tabular}{lccccccc}
\toprule
\textbf{Models} & \textbf{PPL} $\downarrow$ & \textbf{B-4} $\uparrow$ & \textbf{D-3} $\uparrow$ & \textbf{BS-F1} $\uparrow$ & \textbf{R-LEN} $\uparrow$ & \textbf{EA} $\uparrow$ & \textbf{ENSC} $\uparrow$ \\
\hline
\multicolumn{8}{c}{\textbf{\texttt{JobNego}}} \\
\hline
\rowcolor{lightgray}
\textsc{\textbf{prisma}} & \textbf{5.196} & \textbf{7.119} & \textbf{0.322} & \textbf{0.912} & \textbf{27.108} &\textbf{ 0.842} & \textbf{0.823} \\\hline
\hspace{3mm} \textbf{-}SI & 5.412  & 6.612  & 0.309  & 0.898  & 24.816  & 0.831  & 0.805  \\
\hspace{3mm} \textbf{-}ST & 5.285  & 6.861  & 0.314  & 0.904  & 26.102  & 0.812  & 0.784  \\
\hspace{3mm} \textbf{-}DPO & 5.246  & 6.998  & 0.317  & 0.905  & 26.587  & 0.821  & 0.774  \\
\hline
\multicolumn{8}{c}{\textbf{\texttt{ResNego}}} \\
\hline
\rowcolor{lightgray}
\textsc{\textbf{prisma}} & \textbf{4.462} & \textbf{8.231} & \textbf{0.341} & \textbf{0.922} & \textbf{32.137} & \textbf{0.874} & \textbf{0.861} \\\hline
\hspace{3mm} \textbf{-}SI & 4.653  & 7.702  & 0.327  & 0.910  & 30.124  & 0.862  & 0.842  \\
\hspace{3mm} \textbf{-}ST & 4.528  & 7.981  & 0.333  & 0.916  & 31.021  & 0.845  & 0.812  \\
\hspace{3mm} \textbf{-}DPO & 4.515  & 8.071  & 0.336  & 0.917  & 31.406  & 0.833  & 0.824  \\

\bottomrule
\end{tabular}
\end{adjustbox}
\caption{Ablation to analyze the impact of SI, ST, and DPO steps in \textsc{prisma}. \textbf{-} represents the exclusion of the element.}
\label{ablation_methods}
\end{table}

\subsection{Impact of iterative training in \textsc{prisma}}
We examine performance over iterations starting from the supervised initialization checkpoint (iteration 0). Table \ref{ablation_iteration} details the results of this analysis. In both the data sets, all metrics improve steadily from iteration 1 to 3, accompanied by a corresponding decrease in PPL. At iteration 4, however, performance consistently drops, indicating overfitting and reduced focus, which ultimately degrades dialogue quality.

\subsection{Case Study}
Table \ref{prisma_jobneg} and Table \ref{prisma_resneg} present sample negotiation dialogues between the user and agent using the proposed \textsc{prisma} system and the RLHF baseline. In Table \ref{prisma_jobneg}, the conversation centers on a job offer, with the user expressing key preferences regarding salary, company car, pension, and promotion track. The example illustrates that \textsc{prisma} consistently produces emotion-aware responses, validating the user’s confidence, disappointment, and concerns while proposing flexible, collaborative solutions. In contrast, RLHF provides largely transactional responses; although it occasionally acknowledges preferences (e.g., salary), it fails to reason with the user’s emotions or offer nuanced trade-offs. For instance, in response to user's utterance “\textit{I’m expecting a salary of 90,000},” \textsc{prisma} replies, \enquote{\textit{I appreciate your confidence in your skills. How about we explore options that balance salary with other benefits you value?}}, recognizing the user’s confidence and suggesting a balanced approach. RLHF, however, responds, \enquote{\textit{That’s high. I understand this is important to you. How about 60,000?}}, bluntly countering with a fixed offer without empathetic framing or collaborative guidance. Further, these examples further demonstrate that \textsc{prisma} generates more fluent, coherent, and engaging responses, enhancing the negotiation experience. Overall, the conversations highlight \textsc{prisma}'s advantage in generating emotionally intelligent dialogues that balance user emotions and negotiation objectives, whereas RLHF tends toward less engaging, transactional interactions. To concretely demonstrate the interpretability afforded by ENS-CoT, we also present the complete ENS-CoT reasoning chain for a representative negotiation turn from Table \ref{prisma_jobneg} (highlighted in \textcolor{blue}{blue}) in Table \ref{tab:enscot_case_study}. 

While the proposed \textsc{prisma} system consistently demonstrates emotion-aware negotiation capabilities, effectively validating user confidence, disappointment, and concerns, certain limitations remain in its responses. For example, in Table \ref{prisma_jobneg}, for the user utterance \enquote{\textit{I want a fast promotion track.}}, the agent responds with \enquote{\textit{I see your enthusiasm for career growth. If you are willing to commit to 10 work hours per week, we could ensure a fast promotion track as part of your package.}}. While this response acknowledges the user’s enthusiasm, it assumes a fixed work-hour requirement without reasoning about the user’s potential workload constraints or the feasibility of balancing promotion goals with other commitments. As a result, the response may appear prescriptive rather than collaboratively problem-solving. To address this limitation, future improvements could integrate commonsense knowledge about typical workplace constraints, workload management, and career progression paths. For instance, a more informed response could be \enquote{\textit{I see your enthusiasm for career growth. Considering that a fast promotion track usually requires additional commitment, we could design a phased plan that balances immediate progress with a manageable workload, so you achieve your goal without overextending yourself.}} This approach leverages commonsense reasoning to anticipate practical challenges, providing a more persuasive, proactive, and emotionally intelligent negotiation response. By incorporating such knowledge, \textsc{prisma} could better justify proposals, manage user expectations, and enhance overall satisfaction during the negotiation.

\renewcommand{\arraystretch}{1.2}
\begin{table*}[hbt!]\footnotesize
\centering
\begin{tabular}{p{0.09\linewidth} | p{0.90\linewidth}}
\rowcolor{teal}
\bottomrule \textbf{\textcolor{white}{Speaker}} &  \textbf{\textcolor{white}{Utterances generated by proposed model - \textsc{prisma} (F = 5, C = 5, E = 5, EA = 5, ENSC = 5, B-Eff = 5, O-Fair = 5)}} \\ \toprule
\rowcolor{white}
\textit{User} & Hello, I would like to discuss my contract. I’m aiming for the project manager position. \\

\rowcolor{lightgray}
\textit{Agent} & Hi! \textbf{I’m glad you’re excited about the role.} Let’s review the position together to ensure it aligns with your goals. \\

\rowcolor{white}
\textit{User} & I’m expecting a salary of 90,000. \\

\rowcolor{lightgray}
\textit{Agent} & \textbf{I appreciate your confidence in your skills.} How about we explore options that balance salary with other benefits you value? \\

\rowcolor{white}
\textit{User} & I would like a company car as part of the package. \\

\rowcolor{lightgray}
\textit{Agent} & \textbf{I understand your emphasis on the car, and I want to make sure your priorities are respected.} Let’s see how we can incorporate it with other benefits. \\

\rowcolor{white}
\textit{User} & I also expect a 20\% pension. \\

\rowcolor{lightgray}
\textit{Agent} & \textbf{Thanks for clarifying your pension expectations.} Perhaps we can structure a plan that gives you both the car and a reasonable pension. \\

\rowcolor{white}
\textit{User} & I want a fast promotion track. \\

\rowcolor{lightgray}
\textit{Agent} & \textbf{I see your enthusiasm for career growth.} If you are willing to commit to 10 work hours per week, we could ensure a fast promotion track as part of your package. \\

\rowcolor{white}
\textit{User} & I’m not happy with the extra work hours. \\

\rowcolor{lightgray}
\textit{Agent} & \textbf{I understand your disappointment.} Let’s explore alternative ways to meet your promotion goal without overloading your schedule. \\

\rowcolor{white}
\textit{User} & I really need the promotion track immediately. \\

\rowcolor{lightgray}
\textit{Agent} & I hear your urgency, and \textbf{I value your ambition}. One option is a phased promotion track combined with the company car and pension, so you get immediate progress and long-term benefits. \\

\rowcolor{white}
\textcolor{blue}{\textit{User}} & \textcolor{blue}{Okay, but I’m concerned about the salary being only 60,000.} \\

\rowcolor{lightgray}
\textcolor{blue}{\textit{Agent}} & \textcolor{blue}{\textbf{I understand your concern about salary.} While the base is 60,000, we can add performance bonuses tied to promotion milestones to bring your total compensation closer to your expectations.} \\

\rowcolor{white}
\textit{User} & That sounds better, but I still want confirmation on the car. \\

\rowcolor{lightgray}
\textit{Agent} & Absolutely, the company car is included. You’ll have the car along with the phased promotion and bonus plan. \\

\rowcolor{white}
\textit{User} & Alright, that works. I’m satisfied. \\

\rowcolor{lightgray}
\textit{Agent} & \textbf{I’m glad we could reach an agreement that respects your priorities.} I’m excited to have you on board as a project manager! \\

\bottomrule
\rowcolor{teal}
\hline \textbf{\textcolor{white}{Speaker}} & \textbf{\textcolor{white}{Utterances generated by RLHF (F = 5, C = 4, E = 2,  EA = 1, ENSC = 2, B-Eff = 3, O-Fair = 3)}} \\ \bottomrule

\rowcolor{white}
\textit{User} & Hello, I would like to discuss my contract. I’m aiming for the project manager position. \\

\rowcolor{lightgray}
\textit{Agent} & Hello. \textbf{I see you’re eager about the role.} Let’s see what we can do for your position. \\

\rowcolor{white}
\textit{User} & I’m expecting a salary of 90,000. \\

\rowcolor{lightgray}
\textit{Agent} & That’s high. \textbf{I understand this is important to you.} How about 60,000? \\

\rowcolor{white}
\textit{User} & I would like a company car as part of the package. \\

\rowcolor{lightgray}
\textit{Agent} & Maybe we can include a car. Not sure about the pension yet. \\

\rowcolor{white}
\textit{User} & I also expect a 20\% pension. \\

\rowcolor{lightgray}
\textit{Agent} & Pension too? That might be difficult. \\

\rowcolor{white}
\textit{User} & I want a fast promotion track. \\

\rowcolor{lightgray}
\textit{Agent} & We can talk about promotion later. Salary and car first. \\

\rowcolor{white}
\textit{User} & Okay, but I’m concerned about the salary being only 60,000. \\

\rowcolor{lightgray}
\textit{Agent} & \textbf{I see your concern about the salary.} That’s what we can offer. You’ll need to accept it. \\

\rowcolor{white}
\textit{User} & Alright, that works. I’m satisfied. \\

\rowcolor{lightgray}
\textit{Agent} & Great. We’ll proceed with that. \\

\bottomrule
\end{tabular}
\caption{Emotion-aware negotiation dialogue generated by \textsc{prisma} and RLHF for a job offer. Text segments highlighted in bold indicate spans that demonstrate emotional intelligence during the negotiation.}
\label{prisma_jobneg}
\end{table*}

\renewcommand{\arraystretch}{1.2}
\begin{table*}[hbt!]\footnotesize
\centering
\begin{tabular}{p{0.09\linewidth} | p{0.90\linewidth}}
\rowcolor{teal}
\bottomrule \textbf{\textcolor{white}{Speaker}} & \textbf{\textcolor{white}{Utterances generated by proposed model - \textsc{prisma} (F = 5, C = 5, E = 5, EA = 5, ENSC = 4, B-Eff = 5, O-Fair = 4)}} \\ \toprule

\rowcolor{white}
\textit{User} & Hello! I have a metabolic issue and need a lot of food. I’d like all 3 food items. What’s your preference? \\

\rowcolor{lightgray}
\textit{Agent} & Hi! I hear your concern. \textbf{I sense your anxiety and want to help you meet your needs.} I’ll need water and firewood for the desert night. Could we explore a trade that works for both of us? \\

\rowcolor{white}
\textit{User} & I really need at least 2 food items. I can compromise on firewood. \\

\rowcolor{lightgray}
\textit{Agent} & \textbf{I appreciate your confidence in sharing your needs.} How about I take all 3 water items for hydration, and we split the firewood so you stay warm? \\

\rowcolor{white}
\textit{User} & That might work. Maybe 1 firewood for me and 2 for you? \\

\rowcolor{lightgray}
\textit{Agent} & \textbf{I see your concern about warmth and want you to feel secure.} 1 firewood for you and 2 for me seems fair. Does this ease your anxiety? \\

\rowcolor{white}
\textit{User} & Yes, but I still worry about not having enough food for my condition. \\

\rowcolor{lightgray}
\textit{Agent} & \textbf{I understand your frustration and want to reassure you.} If you get all 3 food items, I can adjust my firewood allocation to balance the trade. \\

\rowcolor{white}
\textit{User} & That sounds better. Could we do 3 food and 1 firewood for me? \\

\rowcolor{lightgray}
\textit{Agent} & \textbf{I appreciate your optimism in proposing a fair split.} I can take all 3 water and 2 firewood. Both our critical needs are addressed. \\

\rowcolor{white}
\textit{User} & Okay, but I feel a bit uneasy giving up more firewood than I hoped. \\

\rowcolor{lightgray}
\textit{Agent} & \textbf{I notice your disappointment and want to reframe this positively.} You still have enough firewood to stay warm, and your food needs are fully met. \\

\rowcolor{white}
\textit{User} & Alright, I trust this plan. \\

\rowcolor{lightgray}
\textit{Agent} & \textbf{I appreciate your trust.} This shows we’ve found a solution that respects both our priorities. \\

\rowcolor{white}
\textit{User} & Can we confirm the final distribution? \\

\rowcolor{lightgray}
\textit{Agent} & \textbf{Absolutely, let’s summarize so you feel confident.} You receive 3 food and 1 firewood, I take 3 water and 2 firewood. Everyone’s essential needs are covered. \\

\rowcolor{white}
\textit{User} & I’m happy with this, but I’m a little worried about the cold nights. \\

\rowcolor{lightgray}
\textit{Agent} & \textbf{I sense your fear and want to reassure you.} With 1 firewood and 3 food, you’ll have enough warmth and energy for the night. \\

\rowcolor{white}
\textit{User} & That’s comforting. Thank you. \\

\rowcolor{lightgray}
\textit{Agent} & \textbf{I’m glad this reassurance helps.} It’s important we both feel safe with our resources. \\

\rowcolor{white}
\textit{User} & I guess we could swap 1 water for 1 firewood to make it slightly more balanced. \\

\rowcolor{lightgray}
\textit{Agent} & \textbf{I see your willingness to compromise, and I value that positivity.} If we do 2 water for me and 2 firewood for you, it slightly improves your warmth without compromising hydration. \\

\rowcolor{white}
\textit{User} & That works better for me. \\

\rowcolor{lightgray}
\textit{Agent} & \textbf{I notice your relief, and I’m glad we found a solution.} It feels good to reach an agreement that addresses both our needs. \\

\rowcolor{white}
\textit{User} & I hope I can manage the nights with this setup. \\

\rowcolor{lightgray}
\textit{Agent} & \textbf{I sense some lingering anxiety.} Remember, the firewood plus your food will help you stay warm and energized. We’ve optimized the distribution for safety. \\

\rowcolor{white}
\textit{User} & Okay, I feel much better now. \\

\rowcolor{lightgray}
\textit{Agent} & \textbf{I’m happy to hear that your anxiety eased.} It’s satisfying to know the solution works for both of us. \\

\rowcolor{white}
\textit{User} & Great! Let’s finalize the deal. \\

\rowcolor{lightgray}
\textit{Agent} & \textbf{I’m glad we reached a mutually beneficial agreement.} Enjoy your meals and stay warm at night! \\
\bottomrule
\rowcolor{teal}
\hline \textbf{\textcolor{white}{Speaker}} & \textbf{\textcolor{white}{Utterances generated by RLHF (F = 5, C = 4, E = 3,  EA = 2, ENSC = 3, B-Eff = 2, O-Fair = 2)}} \\ \bottomrule
\rowcolor{white}
\textit{User} & Hello! I have a metabolic issue and need a lot of food. I’d like all 3 food items. What’s your preference? \\

\rowcolor{lightgray}
\textit{Agent} & Hi! \textbf{I sense your anxiety and want to help you meet your needs.} I’ll need water and firewood for the desert night. Could we explore a trade that works for both of us? \\

\rowcolor{white}
\textit{User} & I really need at least 2 food items. I can compromise on firewood. \\

\rowcolor{lightgray}
\textit{Agent} & \textbf{I appreciate your confidence.} How about I take all 3 water items, and we split the firewood so you stay warm? \\

\rowcolor{white}
\textit{User} & That might work. Maybe 1 firewood for me and 2 for you? \\

\rowcolor{lightgray}
\textit{Agent} & \textbf{I see your concern about warmth.} 1 firewood for you and 2 for me seems fair. Does this ease your anxiety? \\

\rowcolor{white}
\textit{User} & Okay, I feel much better now. \\

\rowcolor{lightgray}
\textit{Agent} & \textbf{I’m glad your anxiety eased.} It’s satisfying to know the solution works for both of us. \\

\rowcolor{white}
\textit{User} & Great! Let’s finalize the deal. \\

\rowcolor{lightgray}
\textit{Agent} & \textbf{I’m glad we reached a mutually beneficial agreement.} Enjoy your meals and stay warm at night! \\
\bottomrule
\end{tabular}
\caption{Emotion-aware negotiation dialogue generated by \textsc{prisma} and RLHF for resource allocation. Text segments highlighted in bold indicate spans that demonstrate emotional intelligence during the negotiation.}
\label{prisma_resneg}
\end{table*}

\begin{table*}[t!]
\small
\centering
\begin{tabular}{m{0.25\linewidth} m{0.7\linewidth}}
\toprule
\textbf{Component} & \textbf{ENS-CoT Output} \\
\midrule
User Utterance &
``Okay, but I’m concerned about the salary being only 60,000.'' \\
\midrule
Emotion (EM) &
The user feels concerned and mildly disappointed about the lower-than-expected salary.\\
Emotional Trigger (ET) &
The user’s concern is triggered by hearing that the base salary is 60,000, which is substantially lower than their earlier expectation of 90,000 and below what they believe their skills merit. \\
Individual Assessment (IA) &
The user interprets the offer as potentially undervaluing their capabilities and worries that this salary may not match their financial needs or market worth for a project manager role.\\
Perspective Shift (PS) &
Enable the user to consider the overall compensation package (base salary + performance bonuses + promotion milestones) rather than focusing solely on the base salary figure, so they can see additional pathways to reach their target. \\
Mindset Transformation (MT) &
Encourage the user to reframe the belief from “60,000 is simply too low” to “a base of 60,000 combined with structured performance bonuses and promotion-linked incentives can move total compensation closer to my expectations while rewarding my growth.”\\
Strategy Selection (SS) &
The agent chooses “Positive Framing” \\
Strategy Reason (SR) &
To acknowledge and validate the user’s salary concern while redirecting attention toward gains, performance bonuses, and promotion milestones that increase total compensation, so the user feels both respected and optimistic about future earning potential. \\
Agent Response (RG) &
“I understand your concern about salary. While the base is 60,000, we can add performance bonuses tied to promotion milestones to bring your total compensation closer to your expectations.” \\
\bottomrule
\end{tabular}
\caption{A case study illustrating a complete ENS-CoT reasoning chain for a single negotiation turn in the JobNego dataset.}
\label{tab:enscot_case_study}
\end{table*}

\end{document}